\documentclass[runningheads]{llncs}
\usepackage[T1]{fontenc}
\usepackage{graphicx}
\usepackage[export]{adjustbox}
\usepackage{tabularx}
\usepackage{pdflscape}
\usepackage{geometry}

\usepackage{amsmath}
\usepackage{amssymb}
\usepackage{amsfonts}

\usepackage{caption}
\usepackage{subcaption}
\usepackage{hyperref}

\begin{document}
\title{How to turn your camera into a perfect pinhole model}
\titlerunning{How to turn your camera into a perfect pinhole model}
%
\author{Ivan De Boi\inst{1}\orcidID{0000-0003-3060-262X} \and
Stuti Pathak\inst{1}\orcidID{0009-0001-8973-4822} \and
Marina Oliveira\inst{2}\orcidID{0000-0001-9271-0357} \and
Rudi Penne\inst{1}\orcidID{0000-0002-0921-1950}}
\authorrunning{De Boi et al.}
%

\institute{InViLab, University of Antwerp, Groenenborgerlaan 171, 2020 Antwerp, Belgium\\
\email{ivan.deboi@uantwerpen.be}\\
\url{http://www.invilab.be}\\
\and{Institute of Systems and Robotics, Department of Electrical and Computer Engineering, University of Coimbra, 3004-531 Coimbra, Portugal}
}

\maketitle              
\begin{abstract} 
Camera calibration is a first and fundamental step in various computer vision applications. Despite being an active field of research, Zhang's method remains widely used for camera calibration due to its implementation in popular toolboxes like MATLAB and OpenCV. However, this method initially assumes a pinhole model with oversimplified distortion models. In this work, we propose a novel approach that involves a pre-processing step to remove distortions from images by means of Gaussian processes. 

Our method does not need to assume any distortion model and can be applied to severely warped images, even in the case of multiple distortion sources, e.g., a fisheye image of a curved mirror reflection. The Gaussian processes capture all distortions and camera imperfections, resulting in virtual images as though taken by an ideal pinhole camera with square pixels. Furthermore, this ideal GP-camera only needs one image of a square grid calibration pattern.

This model allows for a serious upgrade of many algorithms and applications that are designed in a pure projective geometry setting but with a performance that is very sensitive to non-linear lens distortions. We demonstrate the effectiveness of our method by simplifying Zhang's calibration method, reducing the number of parameters and getting rid of the distortion parameters and iterative optimization. We validate by means of synthetic data and real world images. The contributions of this work include the construction of a virtual ideal pinhole camera using Gaussian processes, a simplified calibration method and lens distortion removal. 

\end{abstract}
\section{Introduction}

Camera calibration is a vital first step in numerous computer vision applications, ranging from photogrammetry \cite{li2022review} and depth estimation \cite{mertan2022single} to robotics \cite{khan2019vision} and SLAM \cite{smith2006real,devernay2001straight}. As such, it is still a very active field of research, resulting in a myriad of calibration techniques.  \cite{beardsley1992camera,caprile1990using,raza2019artificial,zheng2022image,puig2012calibration}. In this work, we also use the term \textit{camera} for systems such as multi-camera systems or catadioptric systems \cite{galan2019design} which also include a mirror.

Several attempts have been made to unify the calibration procedures for different camera types and camera systems \cite{Ramalingam2017,Hartley2004,liao2023deep}. However, the most popular method in practice today is still Zhang's method \cite{zhang2000flexible}, which is the basis for the camera calibration toolboxes of both MATLAB and OpenCV. This method assumes a pinhole model with additional lens distortions. The resulting calibration is a compromise between all intrinsic and extrinsic values, including the distortion model parameters. This approach has several drawbacks. First, the pinhole assumption is not applicable to non-central cameras. Second, the calibration is the result of a converging optimization process in which one parameter is adjusted in favour of another one to obtain a better optimum, without actual justification. Third, the proposed distortion models for radial and tangential distortion are in some cases oversimplifications, for instance when the distortion is not perfectly radially symmetric or when the centre of distortion is not at the principle point of the camera. 

In this work, we propose a new approach in which we first perform a pre-processing step on the images to remove all distortions. Next, the undistorted images serve as input for a simplified version of Zhang's method for perfect pinhole cameras. By distortions, we mean all deformations resulting from lenses, camera hardware imperfections, faults in the calibration board and even noise. To capture these, we rely on Gaussian processes \cite{Rasmussen2006}. They are a non-parametric Bayesian regression technique that are very well suited to handle sparse noisy datasets.

The proposed method applies to a variety of 2D-cameras. For any such camera, we train a Gaussian process on the relationship from pixel coordinates of the corners detected in an image of a square grid pattern (e.g., a checkerboard) to a perfectly spread square lattice of virtual 2D points. Only one image of the calibration board is needed for this training. This lattice can be seen as the non-linear projection of the checkerboard corners in the original camera image to a virtual image plane, consisting of virtual pixels. 
The Gaussian process captures all possible distortions. 

All future images can now be mapped to the same virtual image plane. As all distortions are removed, we are left with virtual images as though they were taken by an ideal pinhole camera. This method does not need to assume any distortion model and can be applied to severely warped images, even in the case of multiple distortion sources, e.g., a fisheye image of a curved mirror reflection.

The process of first taking images by the given camera followed by the Gaussian process mapping to this virtual image plane can be considered as acquiring images by a virtual camera, called the \textit{GP-camera}, 
replacing the original camera. We will validate that the imaging by this GP-camera corresponds to a perspectivity (central projective transformation) from the 3D scene to the virtual image plane. In other words, we prove that the GP camera is a pinhole camera. 

The main benefit of obtaining an ideal pinhole camera is that a lot of well-studied algorithms and applications can be employed on its images. These include pose estimation, depth estimation, epipolar geometry, shape from motion, 3D scene reconstruction, optical flow, externally calibrating multiple cameras and other 3D sensors, etc. Many of these assume a central projection. For a treatise on these topics, we refer to the industry standard book \cite{Hartley2004}. Our model allows for a serious upgrade of these algorithms and applications that are designed in a pure projective geometry setting. Their performance is very sensitive to non-linear lens distortions. In particular, the quality of calibration techniques that lean on sphere images is drastically improved when rectified images with square pixels are available \cite{Sun2015,Penne2018,Penne2019}.

In \cite{Ranganathan2012}, Gaussian processes are also used to model lens distortions. However, they serve as a surrogate model for the function that captures the lens distortion. As such, they are still part of the iterative camera calibration process. Lens distortion based on one checkerboard pattern is proposed in \cite{wu2015lens}. However, they implement the Levenberg–Marquardt algorithm to find an optimal set of parameter values for their distortion models. Gaussian processes are non-parametric and as such do not depend on this. A deep learning variant of this can be found in \cite{zhang2022learning}.

The contributions of this works are:
\begin{itemize}
  \item We explain how to construct a virtual ideal pinhole camera out of a variety of non-pinhole cameras using Gaussian processes.
  \item We show that our calibrated GP-camera using a simplified version of Zhang's method leads to more accurate measurements compared to the calibrated original camera using the general Zhang's method with iteration.
  \item We show how our method can be used to remove heavy distortions in images.
\end{itemize}

The rest of the paper is structured as follows: Section \ref{Methods} provides some theoretical background of Gaussian processes. It explains the construction and operation of a virtual GP-camera, and describes how we will validate this pinhole model. In Section \ref{Results} we show the results and compare our method to the MATLAB implementation of Zhang's method. We discuss these results in Section \ref{Discussion}. Finally, we formulate our conclusions in Section \ref{Conclusion}.

\section{Methods}\label{Methods}

\begin{figure}[ht!]
\includegraphics[width=\textwidth]{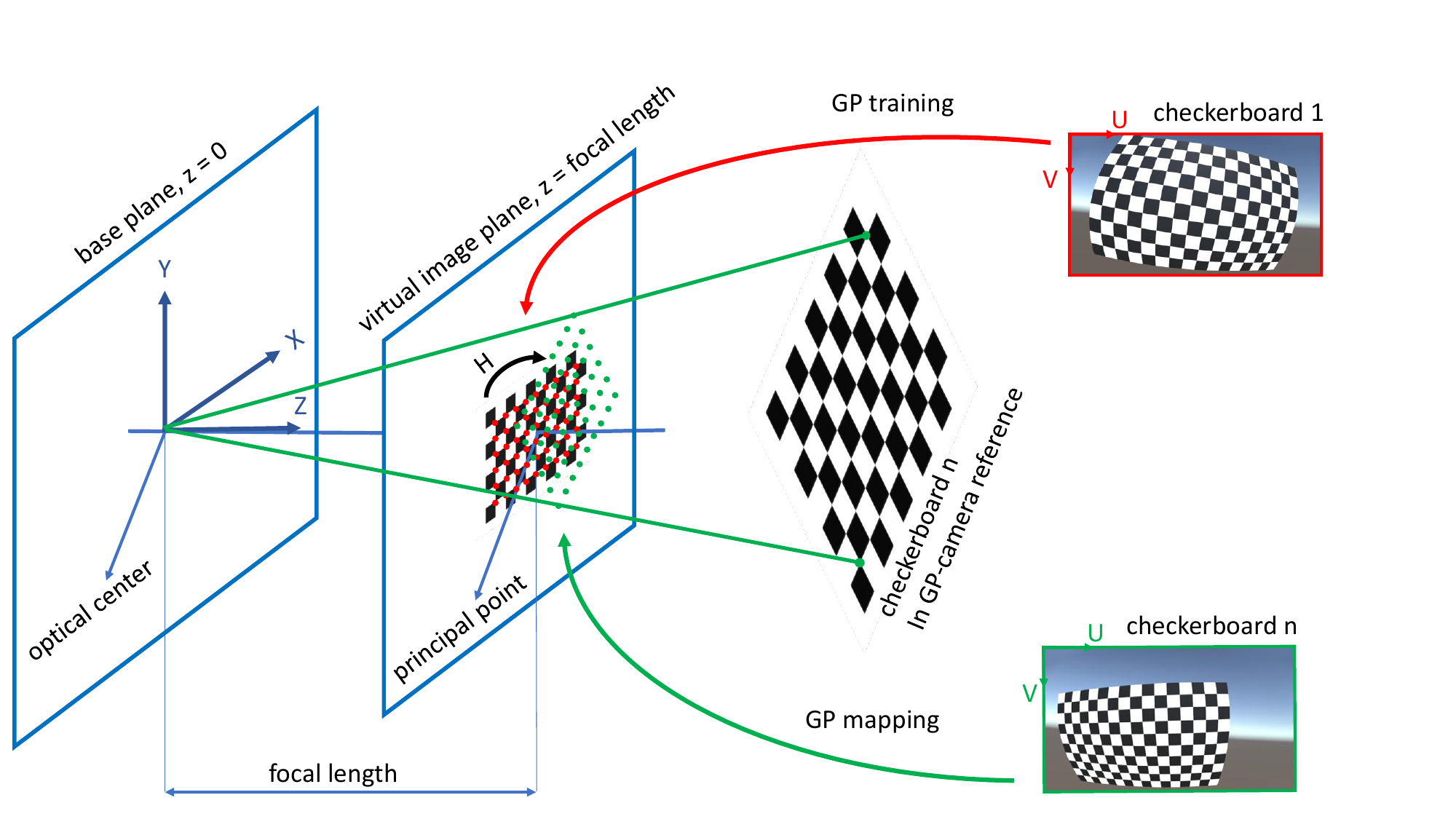}
\caption{An overview of our method. A GP model maps the found corners to the image plane, capturing all lens distortions and camera and checkerboard imperfections.} \label{overview}
\end{figure}

\subsection{Gaussian Processes}

This section includes a brief discussion of the most basic concepts of GPs. As a probabilistic machine learning technique, \textit{Gaussian processes} (GPs) can be employed for numerous prediction tasks, especially in applications where uncertainty estimation plays a key role. They were originally designed as a probabilistic regression method, but appear to be useful as a machine learning technique keeping track of uncertainties.
A Gaussian process can be defined as a continuous collection of random variables, any finite subset of which is normally distributed as a continuous multivariate distribution \cite{Rasmussen2006}. Consider a set of $n$ observations $\mathcal{D} = \{X,\boldsymbol{\mathbf{y}}\}$, where
$X = \begin{bmatrix}
	\boldsymbol{\mathbf{x}}_{1}, \boldsymbol{\mathbf{x}}_{2},..., \boldsymbol{\mathbf{x}}_{n}
\end{bmatrix}^T$
is an $n\times d$ input matrix and
$\boldsymbol{\mathbf{y}} =\begin{bmatrix}
	y_{1}, y_{2},..., y_{n}
\end{bmatrix}^T$
is a $n$-dimensional vector of scalar outputs. We are interested in the mapping $f:\mathbb{R}^d\rightarrow\mathbb{R}$,

\begin{equation}\label{GPR}
	y = f(\boldsymbol{\mathbf{x}})+\epsilon, \quad \epsilon\sim\mathcal{N}(0,\sigma_{\epsilon}^2) ,
\end{equation}
where $\epsilon$ is the identically distributed observation noise.

A GP can be fully defined by its \textit{mean} $m(\boldsymbol{\mathbf{x}})$ and \textit{covariance function} $k(\boldsymbol{\mathbf{x}},\boldsymbol{\mathbf{x}}')$:
\begin{equation}
f(\boldsymbol{\mathbf{x}})\sim\mathcal{GP}(m(\boldsymbol{\mathbf{x}}),k(\boldsymbol{\mathbf{x}},\boldsymbol{\mathbf{x}}'))
\end{equation}
where $\boldsymbol{\mathbf{x}}$ and $\boldsymbol{\mathbf{x}}'$ are two different inputs. It is common practice to normalise the data to zero mean \cite{Rasmussen2006}. By definition, a GP yields a distribution over functions that have a joint normal distribution,	

\begin{equation}\label{joint f f*} 	 	
	\begin{bmatrix}
		\mathbf{f}\\
		\mathbf{f_{*}}
	\end{bmatrix}
	\sim		
	\mathcal{N}		
	\left(
	\begin{bmatrix}
		\boldsymbol{m}_{X}\\
		\boldsymbol{m}_{X_{*}}
	\end{bmatrix}
	,
	\begin{bmatrix}
		\mathbf{K}_{X,X} & \mathbf{K}_{X,X_{*}}\\
		\mathbf{K}_{X_{*},X} & \mathbf{K}_{X_{*},X_{*}}
	\end{bmatrix}
	\right) ,		
\end{equation}
where $X$ and $X_{*}$ are the input vectors of the $n$ observed training points and $n_{*}$ the unobserved test points respectively. 	
The mean value for $\mathbf{X}$ is given by $\boldsymbol{m}_{X}$. Likewise, the mean value for $X_{*}$ is given by $\boldsymbol{m}_{X_{*}}$.
The covariance matrices $\mathbf{K}_{X,X}$, $\mathbf{K}_{X_{*},X_{*}}$, $\mathbf{K}_{X_{*},X}$ and $\mathbf{K}_{X,X_{*}}$ are constructed by evaluating the covariance function $k$ at their respective pairs of points. 
In real world applications, we are depending on noisy observations $\boldsymbol{\mathbf{y}}$, as we don't have access to the latent function values.

The conditional predictive posterior distribution of the GP can be written as:

\begin{equation}\label{pred f f* final}	 	
	\mathbf{f_{*}}|X, X_{*}, \boldsymbol{\mathbf{y}}, \boldsymbol{\theta}, \sigma_{\epsilon}^2 
	\sim		 
	\mathcal{N}\left( \mathbb{E}(\mathbf{f_{*}}), \mathbb{V}(\mathbf{f_{*}}) \right) ,
\end{equation}

\begin{equation}\label{E}	 	
	\mathbb{E}(\mathbf{f_{*}}) = \boldsymbol{m}_{X_{*}} + \mathbf{K}_{X_{*},X}\left[ \mathbf{K}_{X,X}+\sigma_{\epsilon}^2I\right]^{-1}\mathbf{f} ,
\end{equation}

\begin{equation}\label{V}	 	
	\mathbb{V}(\mathbf{f_{*}}) = \mathbf{K}_{X_{*},X_{*}}-\mathbf{K}_{X_{*},X}\left[ \mathbf{K}_{X,X}+\sigma_{\epsilon}^2I\right] ^{-1}\mathbf{K}_{X,X_{*}} .
\end{equation}
In our work, we are mainly interested in $\mathbb{E}(\mathbf{f_{*}})$, the expected value (or mean) of the function values at particular test points. The hyperparameters $\boldsymbol{\theta}$ are usually learned by maximising the log marginal likelihood. In our experiments we use L-BFGS, a quasi-Newton method described in \cite{Liu1989}.
There exists a large variety of covariance functions, also called kernels \cite{KristjansonDuvenaud2014}. A common choice is the squared exponential kernel, which we also employ in our work. It is infinitely differentiable and thus yields smooth functions. This is a reasonable assumption to make in our context and has the following form:
\begin{equation}\label{SE} 	 	
k_{SE}(\mathbf{x}, \mathbf{x}') = \sigma^2_{f}\exp \left( -\frac{\left|\mathbf{x}-\mathbf{x}' \right|^2 }{2l^2}\right) ,
\end{equation}
in which $\sigma^2_{f}$ is a height-scale factor and $l$ the length-scale that determines the radius of influence of the training points. For the squared exponential kernel the hyperparameters are $\sigma^2_{f}$ and $l$.

\subsection{Constructing an ideal pinhole camera}\label{sec:ideal}

Every pixel in an image taken by a camera corresponds to a ray of incoming light, which is a straight line. This means that all points on this straight line in 3D space are mapped to the same pixel. For a perfect \textit{pinhole camera}, as described in \cite{zhang2000flexible}, all these lines intersect in a central point called the optical centre. Consequently, for the pinhole model, taking images corresponds to a perspective projection, which is an important example of a projective map form $\mathbb{P}^3$ to $\mathbb{P}^2$. The image plane is perpendicular to the focal axis, which contains the optical centre and pointing in the direction in which the camera perceives the world. The distance between the optical centre and the image plane is called the focal length. In Figure \ref{overview} we present a pinhole camera, where every checkerboard is projected to the image plane. More details are given in Appendix \ref{App_Zhangs}.

A myriad of calibration techniques for (pinhole) cameras can be found in the literature. Due to its importance in computer vision, this is still a very active field of research. Here, we restrict ourselves to the methods most commonly used in practice, such as the camera calibration toolboxes in MATLAB and OpenCV. Their implementation is based on the well-known Zhang's method \cite{zhang2000flexible}, which is based on the Direct Linear Transform (DLT) method, with the calibration points located in planar objects such as flat checkerboards. See \cite{Hartley2004,burger2016zhang} for a more in depth treatise. We provide an overview of Zhang's method in Appendix \ref{App_Zhangs}. 

To account for distortion, Zhang's method first assumes a perfect pinhole model with no distortion at all, and approximates the calibration matrix $K$ by means of several homographies between checkerboard positions and the image plane. These homographies are used to determine the positions of these boards relative to the camera. These camera intrinsics and extrinsics serve as an initial guess for an iterative process in which the distortion model is integrated. A non-linear optimization process is then implemented to find, after convergence, values for both the intrinsics, extrinsics and the distortion model parameters.

Herein lies the main pitfall of this method. By iterating towards a convergence in the parameters values, there is a compromise between them. This means unjustly altering the value of one parameter in favour of another one. Both the pinhole and the distortion model might be an oversimplification of the underlying reality. 

Now we will explain how we can construct an ideal pinhole camera by using Gaussian processes that first remove all factors that make the pinhole assumption invalid. What is left, is a virtual (ideal) pinhole camera for which a simplified Zhang's method can be used. 

Using a fixed but arbitrary 2D-camera, physical or simulated, we take the image of \textbf{only one} checkerboard, or any planar square grid pattern. We introduce a local $xy$-reference frame on the board plane, with coordinate axes parallel to the grid lines, the unit equal to the square edges and the origin typically coincident with some grid corner. This board can be placed anywhere in 3D space, but for our purposes, it is best to position it in such a way that it fills up the entire image. Once this is done, we define a virtual image plane where the grid squares are the virtual pixels. We detect the grid corners in the image using any corner detection system, e.g., the MATLAB Camera Calibration Toolbox. We assign to every detected corner its local $xy$-coordinates on a virtual image plane. As a consequence, the original board image is mapped to a perfect square lattice of points in the virtual image plane.

These corresponding sets of data are used to train a Gaussian process model for a map between the $uv$-coordinates in the original image plane and the $xy$-coordinates of the virtual image plane, explained in Section \ref{Results}. In practice, we implement two independent scalar output Gaussian processes: one for the resulting $x$-coordinate and one for the $y$-coordinate.

In summary, for a given 2D-camera and the image of some spatially positioned square grid patterns, we have constructed a virtual GP-camera that obtains its images by mapping the real images to the virtual image plane  by means of a Gaussian process. 

Although the Gaussian map of this GP-camera was trained on a single checkerboard image, it apparently removes the distortion for any image of any spatial object.
In Section \ref{Results} we investigate the images for many positions of the calibration board and observe the straightness of all the GP-images of the grid lines.
In other words, the GP-camera maps every plane (board position) as a {\em collineation\/}, which must be a {\em homography\/} according to the fundamental theorem of projective geometry \cite{Sarath1984}. We conclude that the GP-camera images the world as a projective transformation.

Furthermore, this projective transformation is a {\em perspectivity\/} (central projection) since it can be described by the multiplication by
a \textit{projection matrix} $\mathbf{P}$ that can be decomposed as:
\begin{equation}\label{P_decomposed_1} 	 	
\mathbf{x} = \mathbf{P} \mathbf{X} = \mathbf{K} [\mathbf{R}\mid\mathbf{t}] \mathbf{X} ,
\end{equation}
where we work with homogeneous coordinates $\mathbf{x} = (x, y, 1)^T$ for points in the virtual image plane and $\mathbf{X} = (X, Y, Z, 1)^T$ spatial points.

In Section \ref{Results} we determine the calibration matrix $\mathbf{K}$ by a simplified version of Zhang's method, described in Appendix \ref{App_Zhangs_simplified}. It is important to note that we need only three parameters for this. Since we are working on square pixels, there is no longer any skewness and the scale is the same in the $x-$ and $y$-direction:
\begin{equation}\label{K_sim_1} 	 	
\mathbf{K} = \begin{pmatrix}f & 0 & u_c \\ 0 & f & v_c \\ 0 & 0 & 1 \\ \end{pmatrix} .
\end{equation}
The extrinsic component $[\mathbf{R}\mid\mathbf{t}]$ in Equation \ref{P_decomposed_1} transforms the coordinates of the world reference frame to those in the GP-camera reference frame. The GP-origin is determined by the principal point $(u_c,v_c)$ and the focal length $f$, which is measured in virtual pixel units. The $X$ and $Y$ axes of the latter are parallel to the corresponding axes of the virtual image plane as they appear on the checkerboard in its first position. The $Z$ axis is perpendicular to this, yielding the equation $Z=f$ for the GP image plane. 

We validate this pinhole model for the GP-camera by showing the almost perfect match of $\mathbf{K}$ in Equation \ref{K_sim_1} as determined by the equations of Zhang's method for many board positions or homographies. In addition, we observe very small reprojection errors in the pinhole model of the GP-camera, using the relative positions of these boards as provided by the computed homographies (Section \ref{Results}). Everything that makes our real world camera deviate from an ideal pinhole camera is captured by the Gaussian process model. This entails lens distortions, imperfections in the lens, camera, checkerboard and even noise in the image.  

Although the GP-camera is a virtual camera, we have a geometric interpretation of its pinhole model. The centre of this virtual camera coincides with the theoretical optical centre of the physical camera, but the focal axis of the virtual camera is perpendicular to the square lattice board in the position of the first reference image. This lattice board is the virtual image plane, having world pixels equal to the square cells. The focal length is measured from the GP-centre to this virtual image plane in the same grid units. The images of the GP-camera are obtained by a central projection from this centre onto the virtual image plane (Figure \ref{overview}). These dimensions of the virtual camera are fixed once and for all from the moment the first picture is taken. However, they are linked to the physical camera, even if it is later on moved to a different position and rotation.

The calibration of the GP-camera by the simplified Zhang's method not only validates the pinhole model with square pixels, it also provides an interesting alternative to camera calibration, clearly outperforming state-of-the-art methods with respect to simplicity and accuracy (Section \ref{Results}).

\subsection{The datasets}\label{data}

\begin{table}[]
\centering
\caption{The datasets.}
\begin{tabular}{llll}
\hline
Dataset        & Nr of checkerboards & Nr of corners & Image resolution \\ \hline
Unity pinhole                         & 30      &15x9           & 3840x2160        \\
Unity barrel       & 30       &15x9            & 3840x2160        \\
Unity pincushion     & 30       &15x9           & 3840x2160        \\
Webcam & 11        &15x9           & 2560x1440        \\
Webcam with telelens        & 17      &14x9              & 2560x1440              \\
RealSense with mirror        & 26     &14x9               & 971x871              \\ \hline
\end{tabular}
\label{table:T1}
\end{table}

We validate our findings on six diverse datasets of images of checkerboards. An overview is given in Table \ref{table:T1}. The first three datasets are generated in a scene made in the Unity game engine software version (2020.2.5f1). These sets are based on the same scene, so they depict identical positions and rotations for the boards. The barrel distortion and pincushion distortion effect is obtained by the built-in post-processing package. The barrel distortion centre was placed in the centre of the image. The centre of the pincushion distortion is shifted to the left and bottom of the image. 

\begin{figure}[ht!]
     \centering
     \begin{subfigure}[m]{0.23\textwidth}
         \centering
         \includegraphics[width=\textwidth]{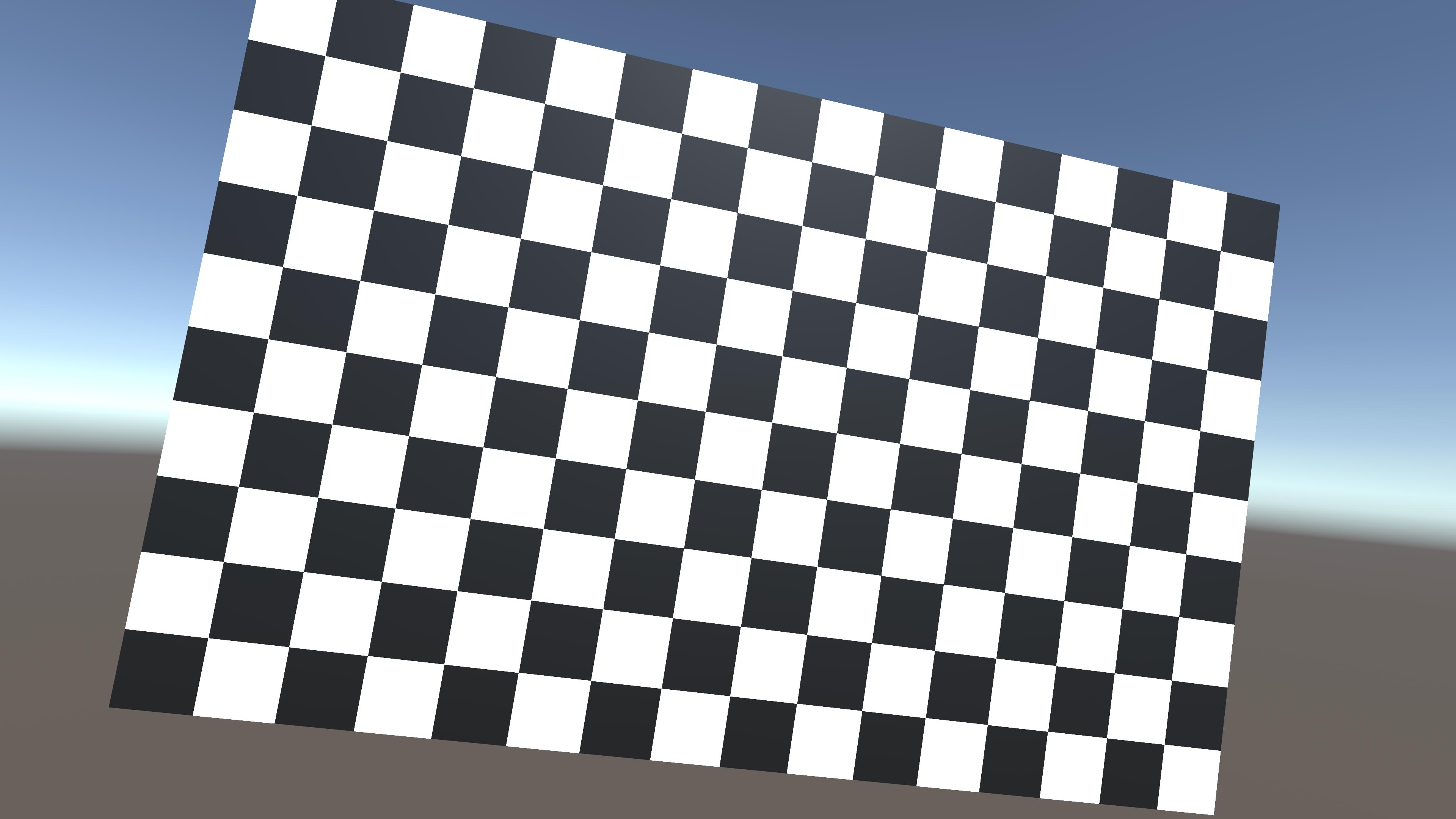}
     \end{subfigure}
     \begin{subfigure}[m]{0.23\textwidth}
         \centering
         \includegraphics[width=\textwidth]{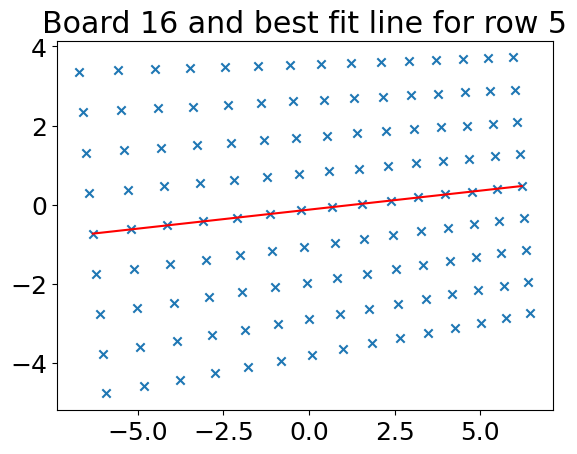}
     \end{subfigure}
     \begin{subfigure}[m]{0.23\textwidth}
         \centering
         \includegraphics[width=\textwidth]{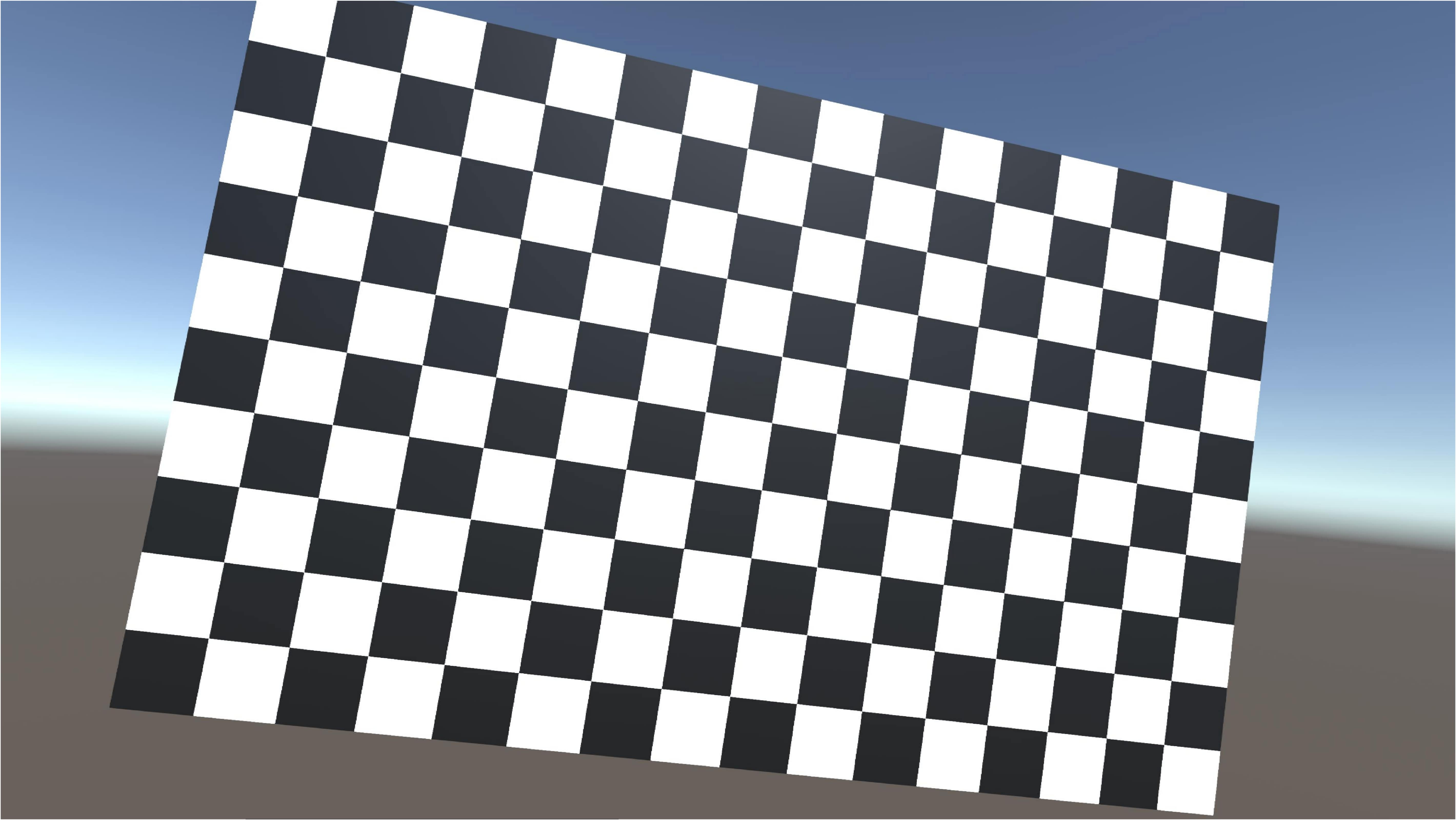}
     \end{subfigure}
    \begin{subfigure}[m]{0.23\textwidth}
    \centering
         \includegraphics[width=\textwidth]{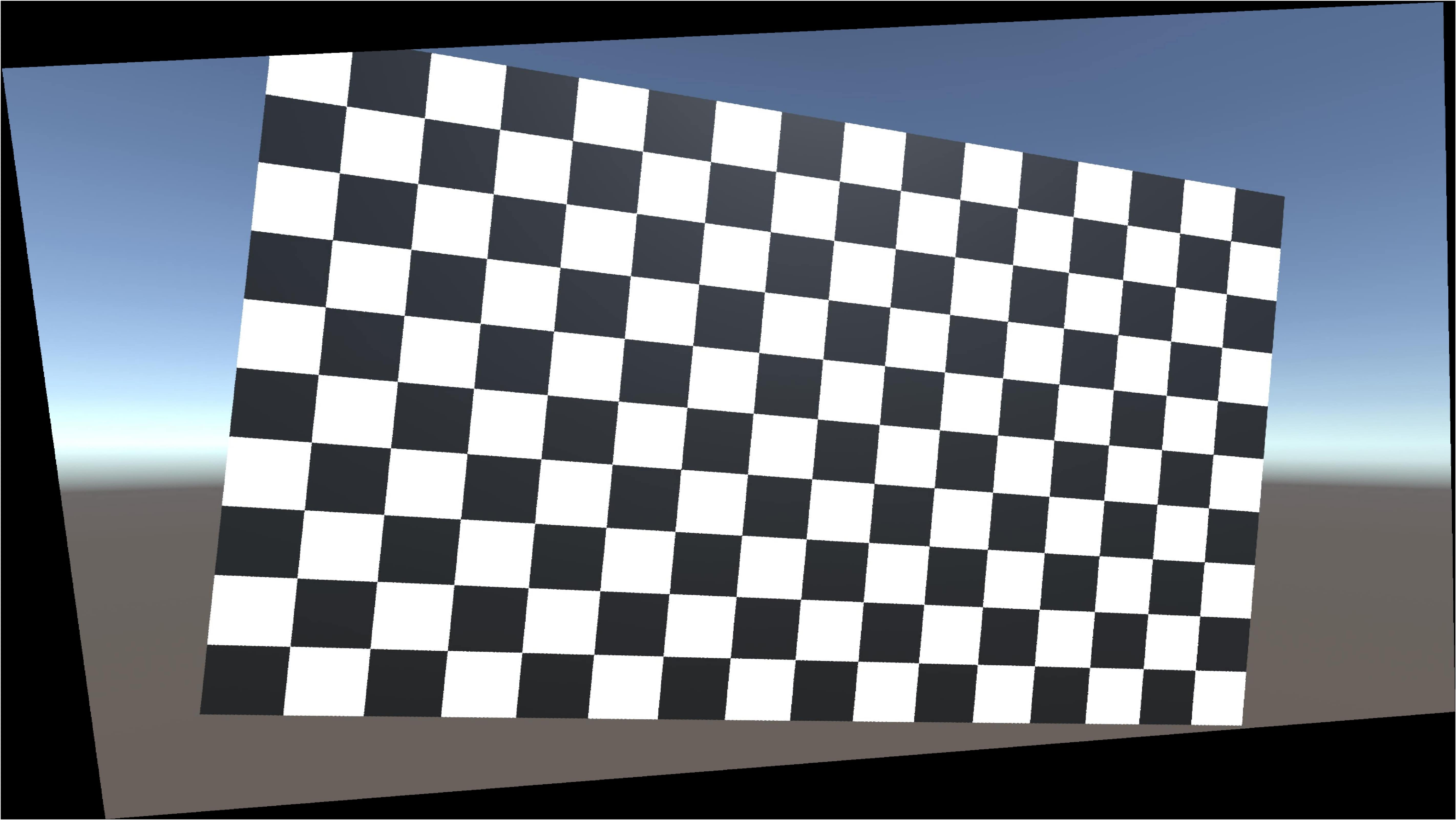}
     \end{subfigure}   
    \vfill
         \begin{subfigure}[m]{0.23\textwidth}
         \centering
         \includegraphics[width=\textwidth]{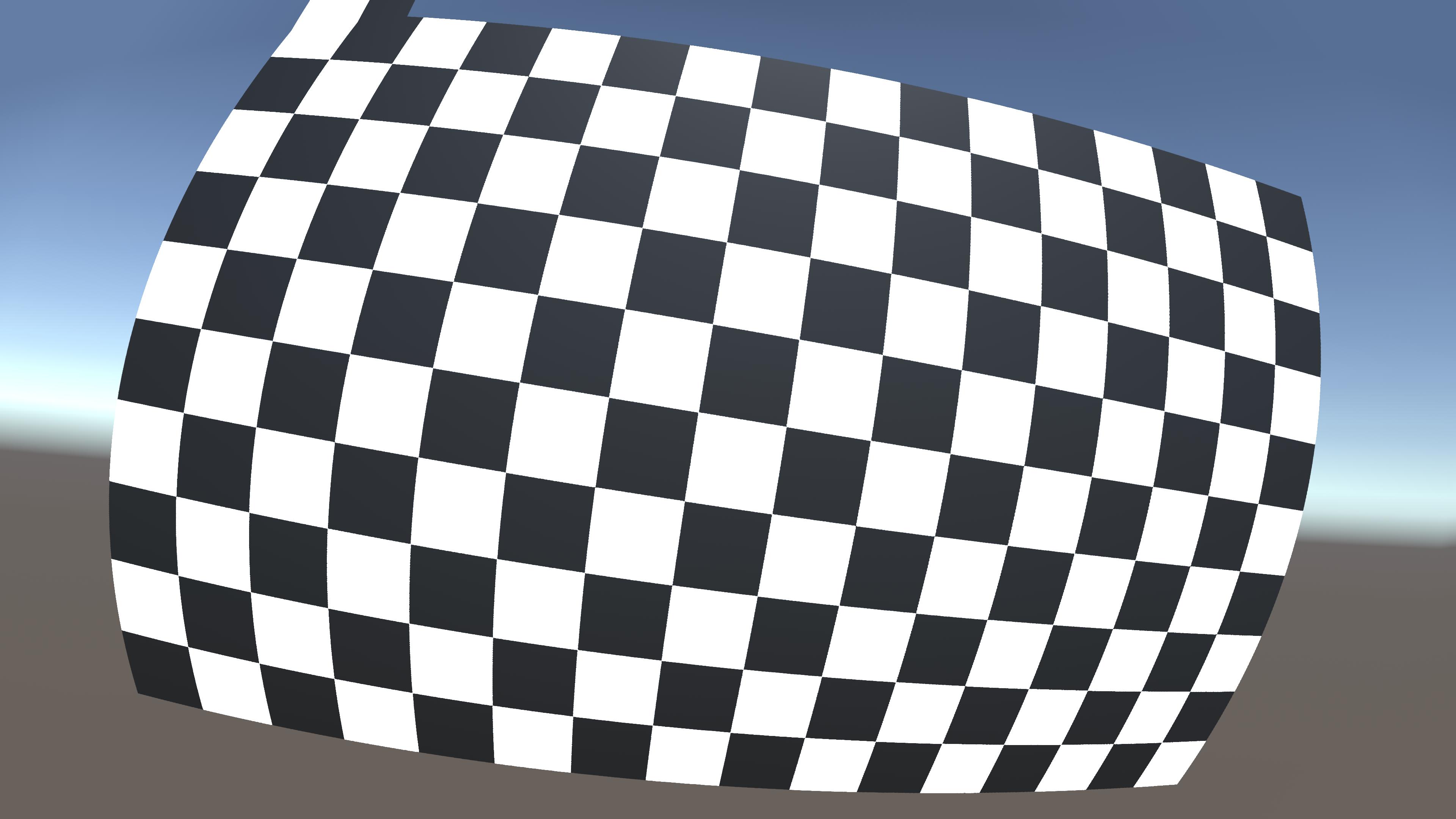}
     \end{subfigure}
     \begin{subfigure}[m]{0.23\textwidth}
         \centering
         \includegraphics[width=\textwidth]{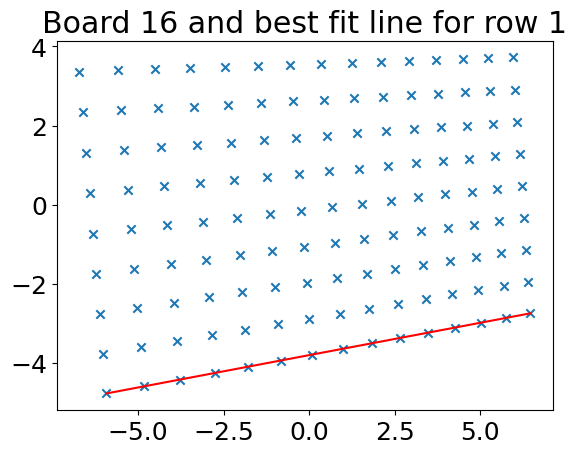}
     \end{subfigure}
     \begin{subfigure}[m]{0.23\textwidth}
         \centering
         \includegraphics[width=\textwidth]{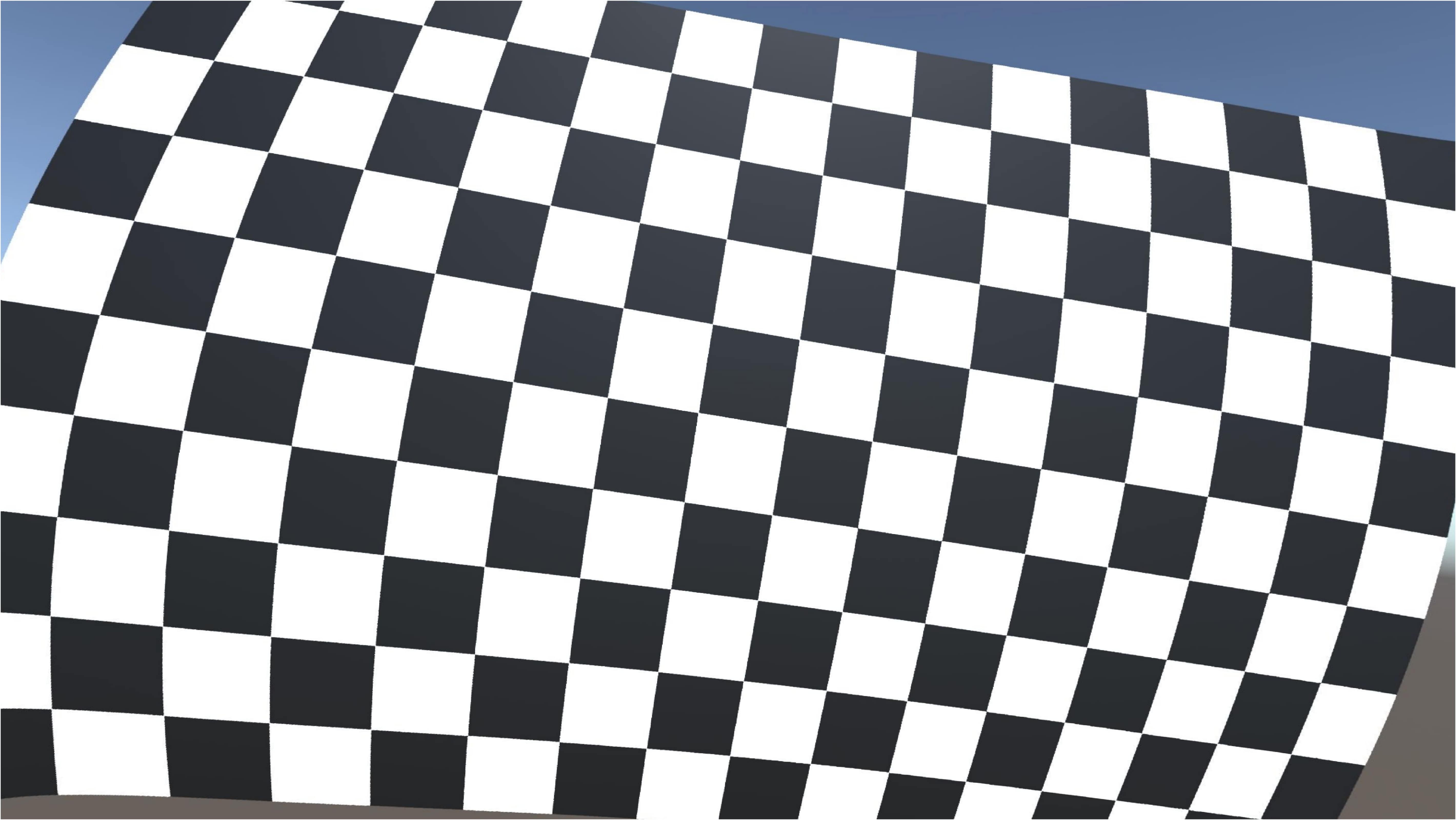}
     \end{subfigure}
    \begin{subfigure}[m]{0.23\textwidth}
    \centering
         \includegraphics[width=\textwidth]{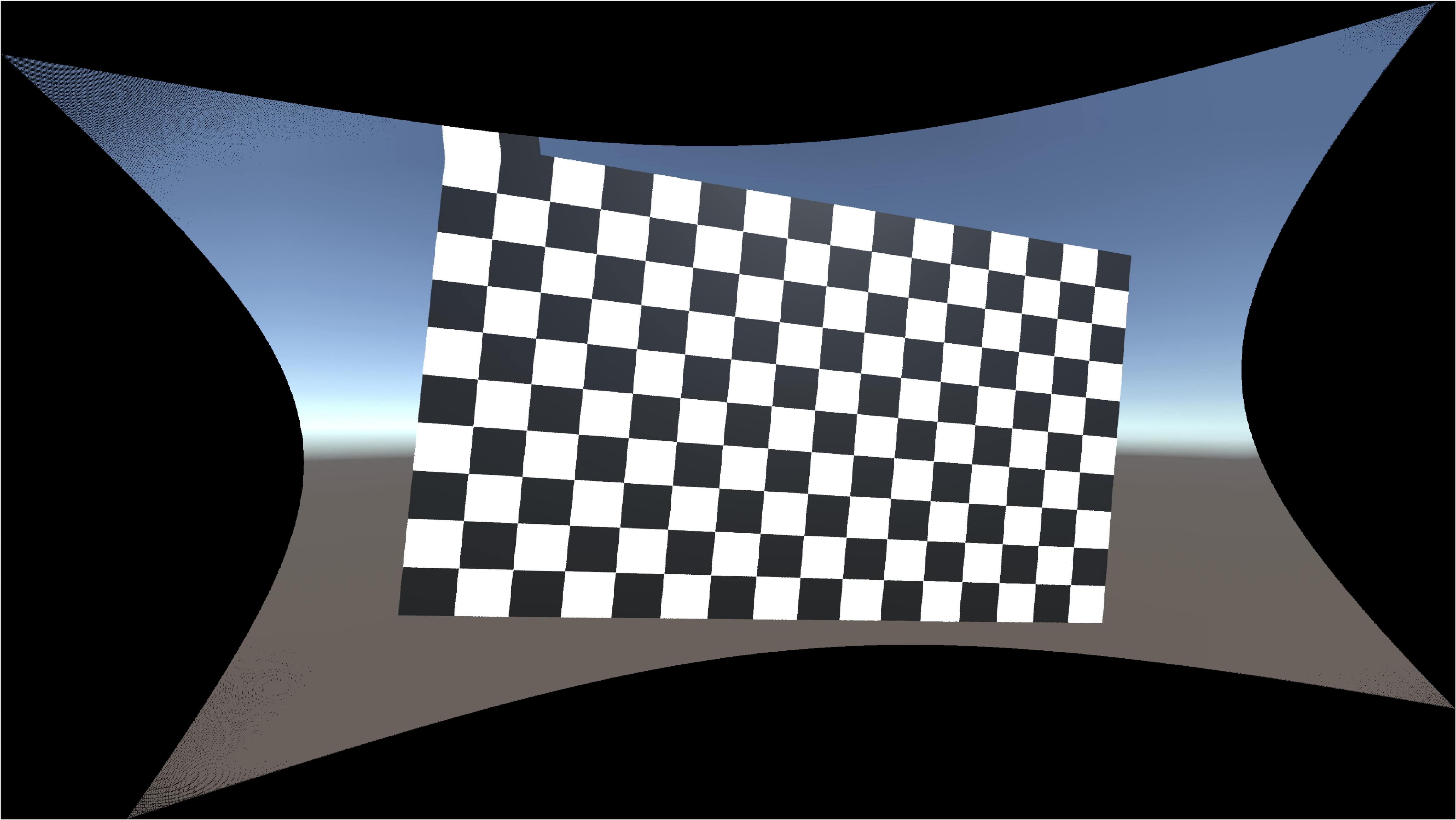}
     \end{subfigure}
    \vfill
         \begin{subfigure}[m]{0.23\textwidth}
         \centering
         \includegraphics[width=\textwidth]{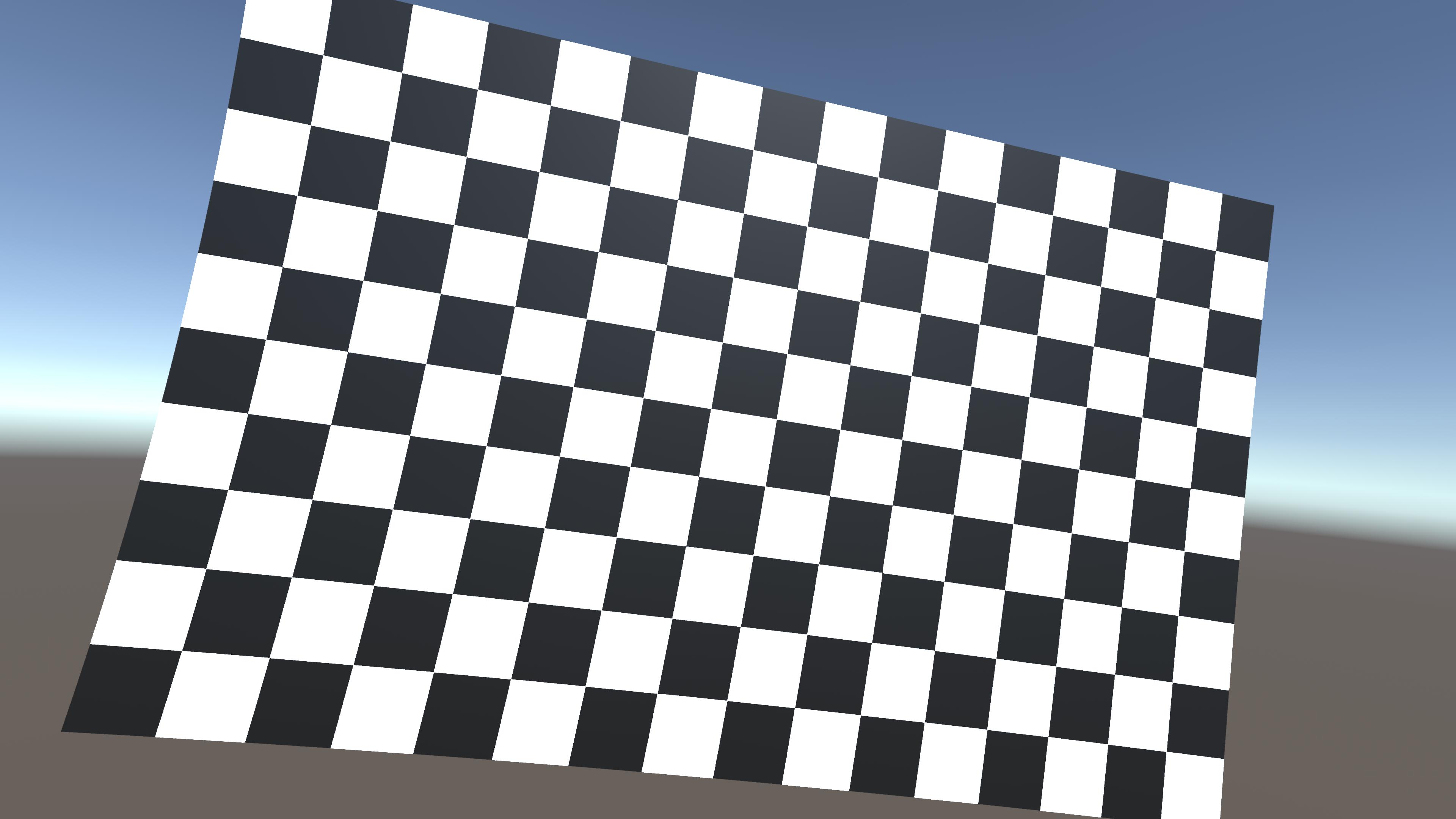}
     \end{subfigure}
     \begin{subfigure}[m]{0.23\textwidth}
         \centering
         \includegraphics[width=\textwidth]{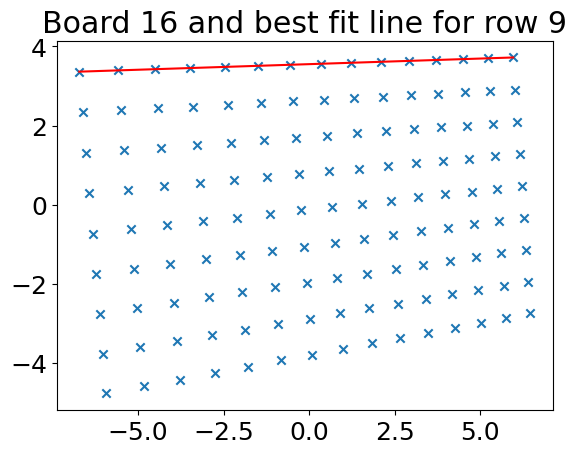}
     \end{subfigure}
     \begin{subfigure}[m]{0.23\textwidth}
         \centering
         \includegraphics[width=\textwidth]{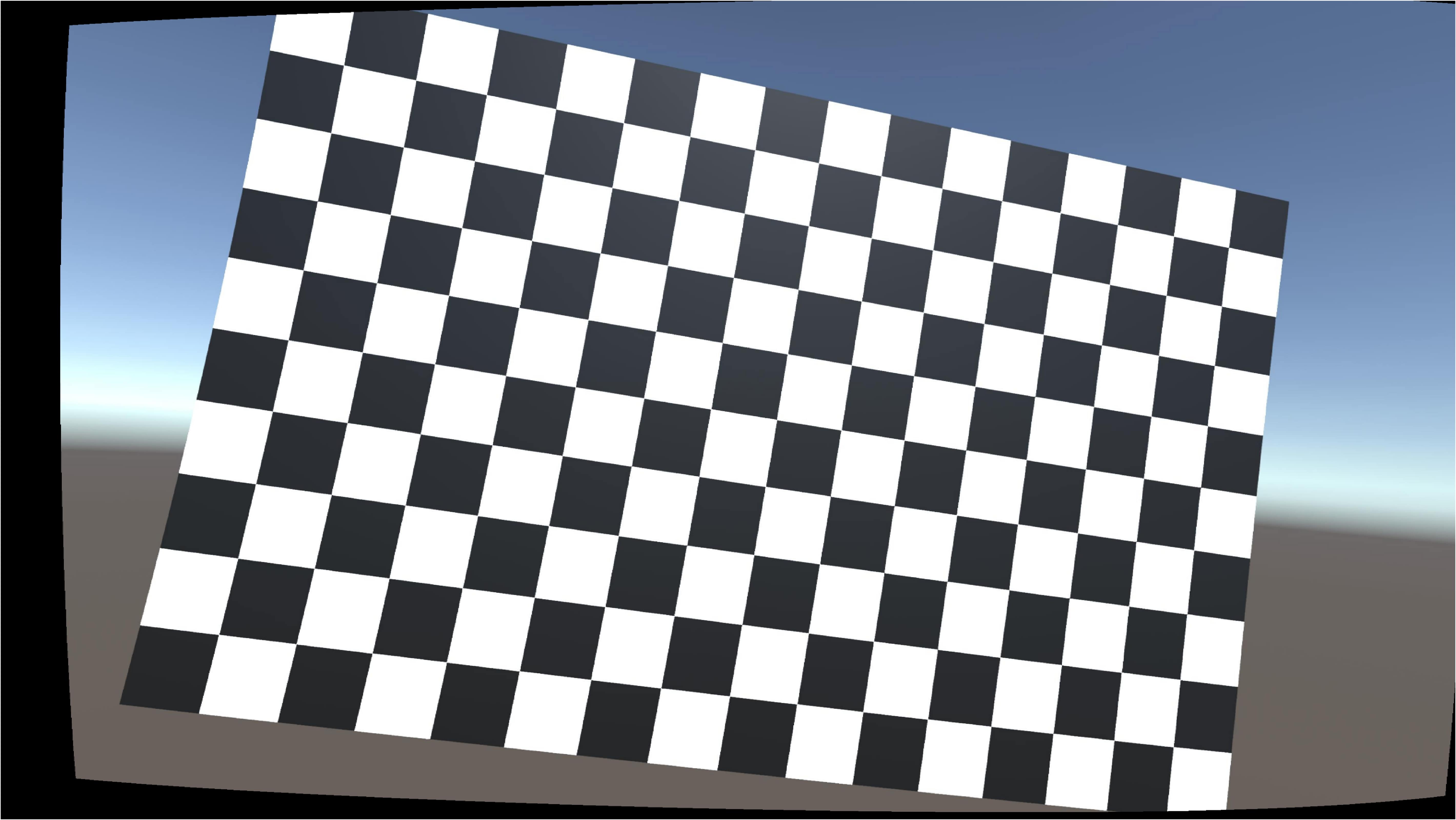}
     \end{subfigure}
    \begin{subfigure}[m]{0.23\textwidth}
    \centering
         \includegraphics[width=\textwidth]{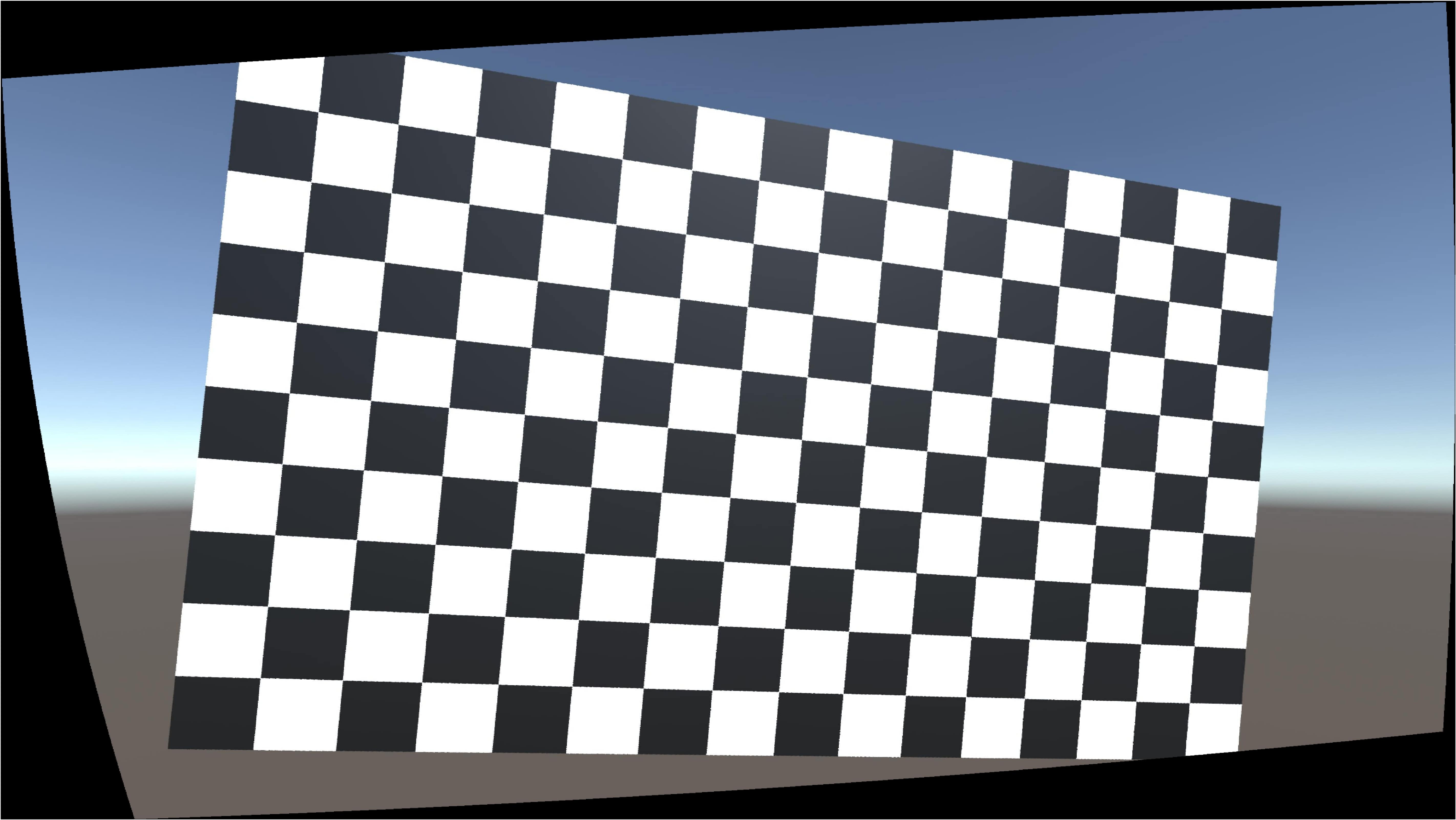}
     \end{subfigure}
         \vfill 
         \begin{subfigure}[m]{0.23\textwidth}
         \centering
         \includegraphics[width=\textwidth]{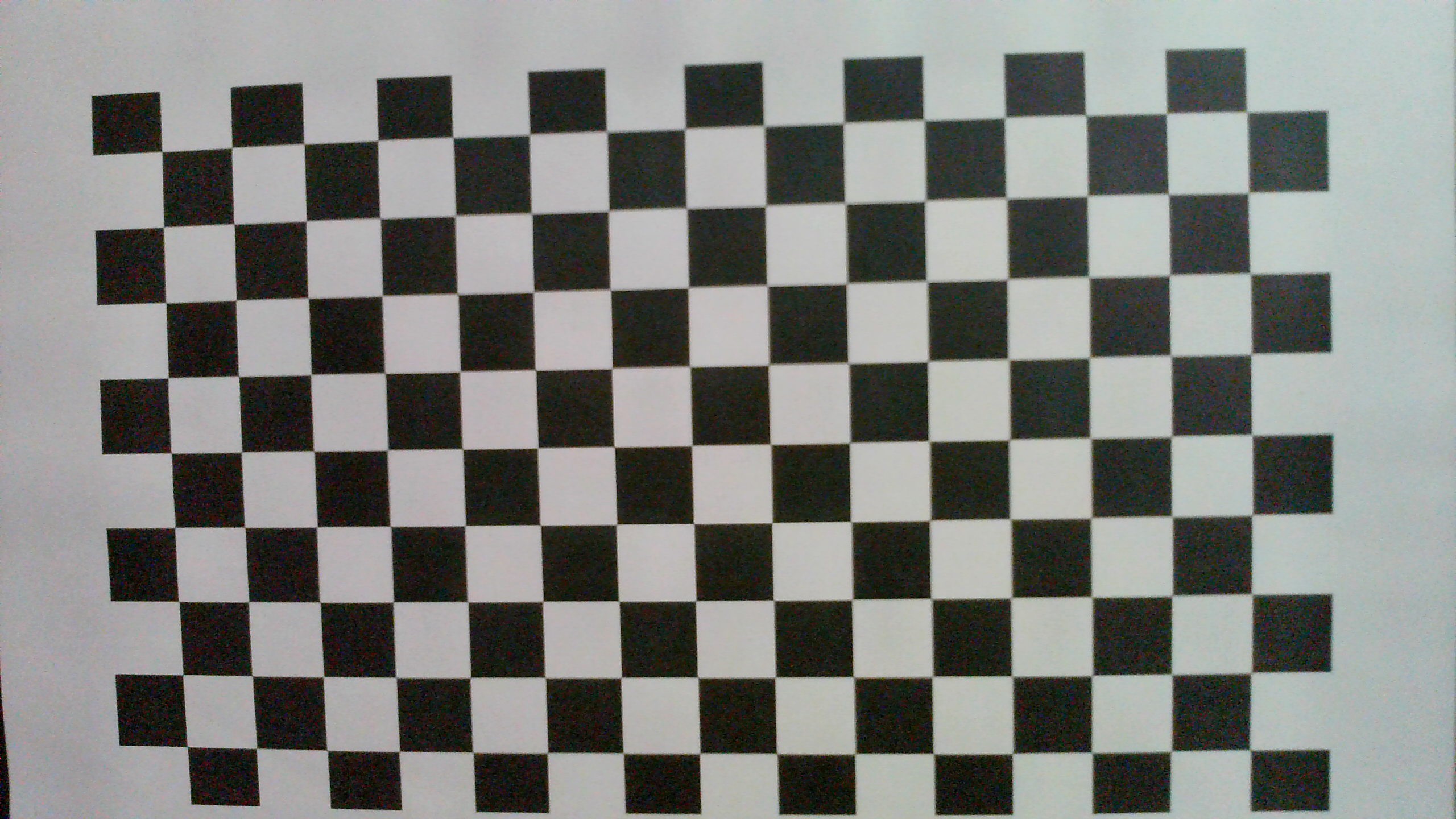}
     \end{subfigure}
     \begin{subfigure}[m]{0.23\textwidth}
         \centering
         \includegraphics[width=\textwidth]{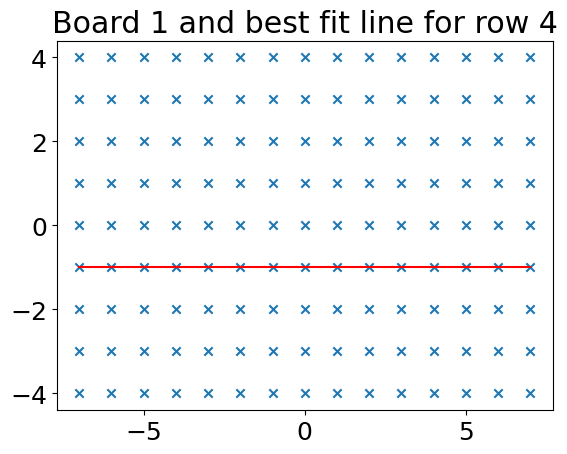}
     \end{subfigure}
     \begin{subfigure}[m]{0.23\textwidth}
         \centering
         \includegraphics[width=\textwidth]{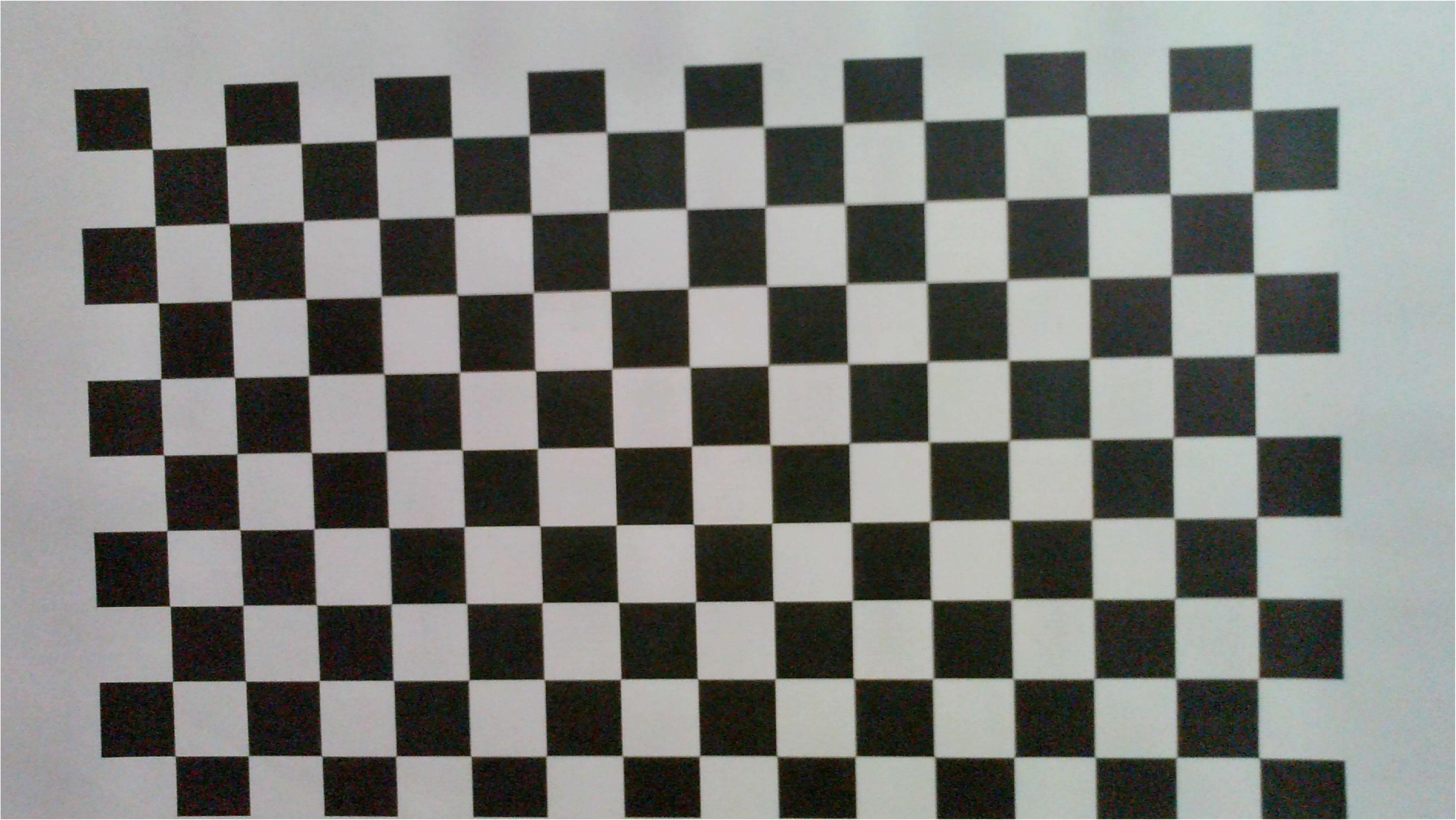}
     \end{subfigure}
    \begin{subfigure}[m]{0.23\textwidth}
    \centering
         \includegraphics[width=\textwidth]{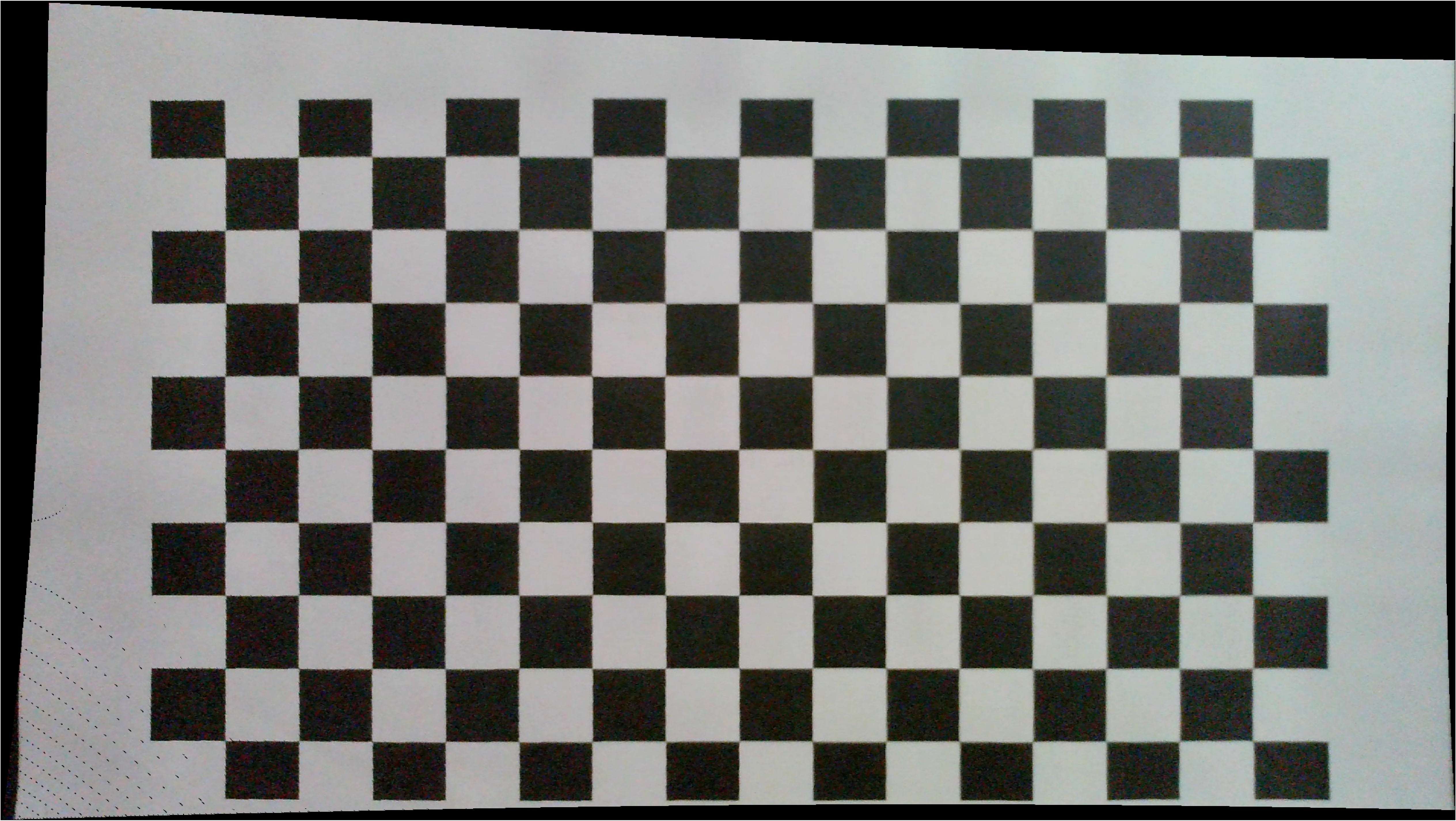}
     \end{subfigure}
         \vfill
         \begin{subfigure}[m]{0.23\textwidth}
         \centering
         \includegraphics[width=\textwidth]{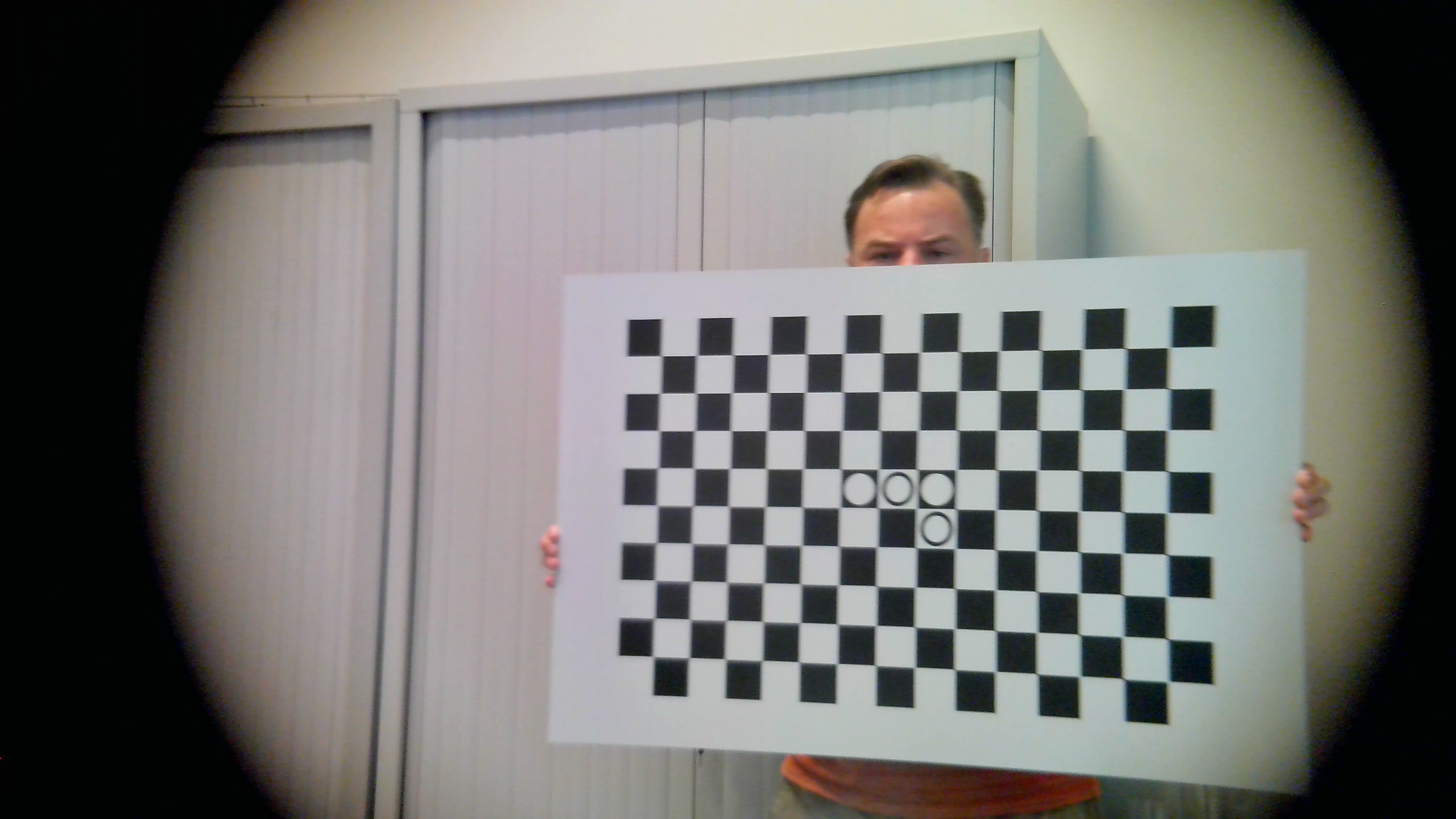}
     \end{subfigure}
     \begin{subfigure}[m]{0.23\textwidth}
     \centering
     \includegraphics[width=\textwidth]{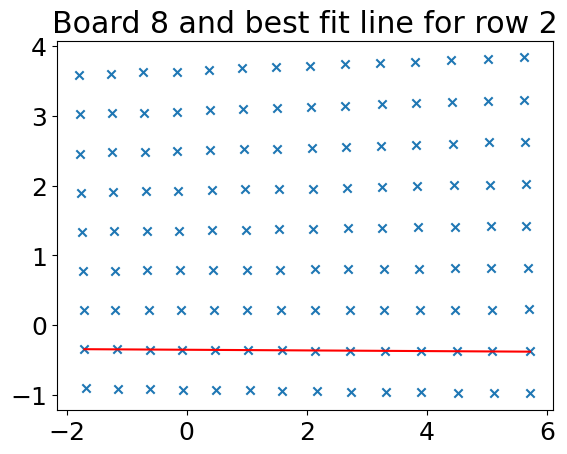}
     \end{subfigure}
     \begin{subfigure}[m]{0.23\textwidth}
     \centering
     \includegraphics[width=\textwidth]{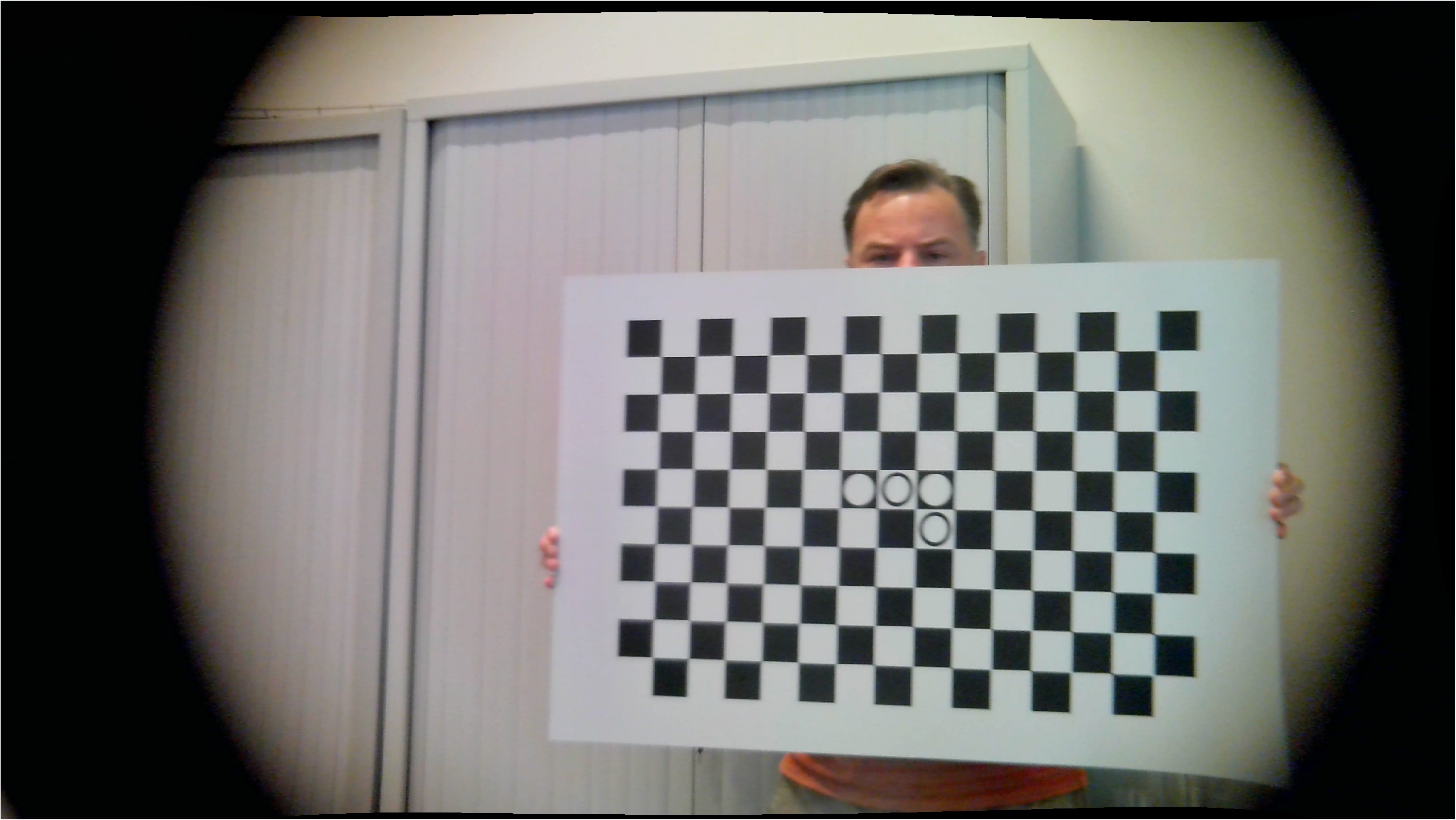}
     \end{subfigure}
    \begin{subfigure}[m]{0.23\textwidth}
    \centering
    \includegraphics[width=\textwidth]{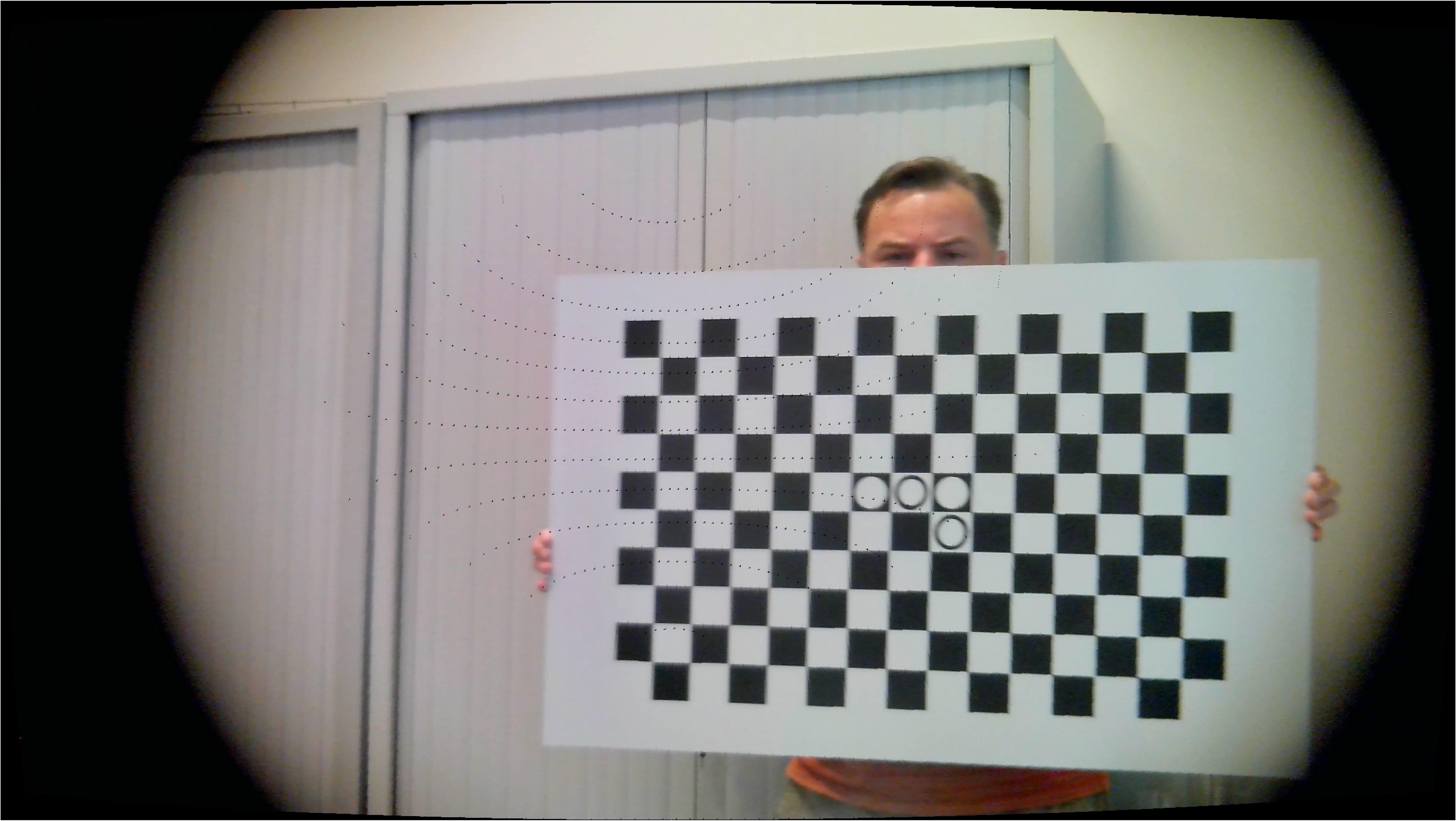}
     \end{subfigure}
     \vfill
     \vfill
     \begin{subfigure}[m]{0.23\textwidth}
     \centering
     \includegraphics[width=\textwidth]{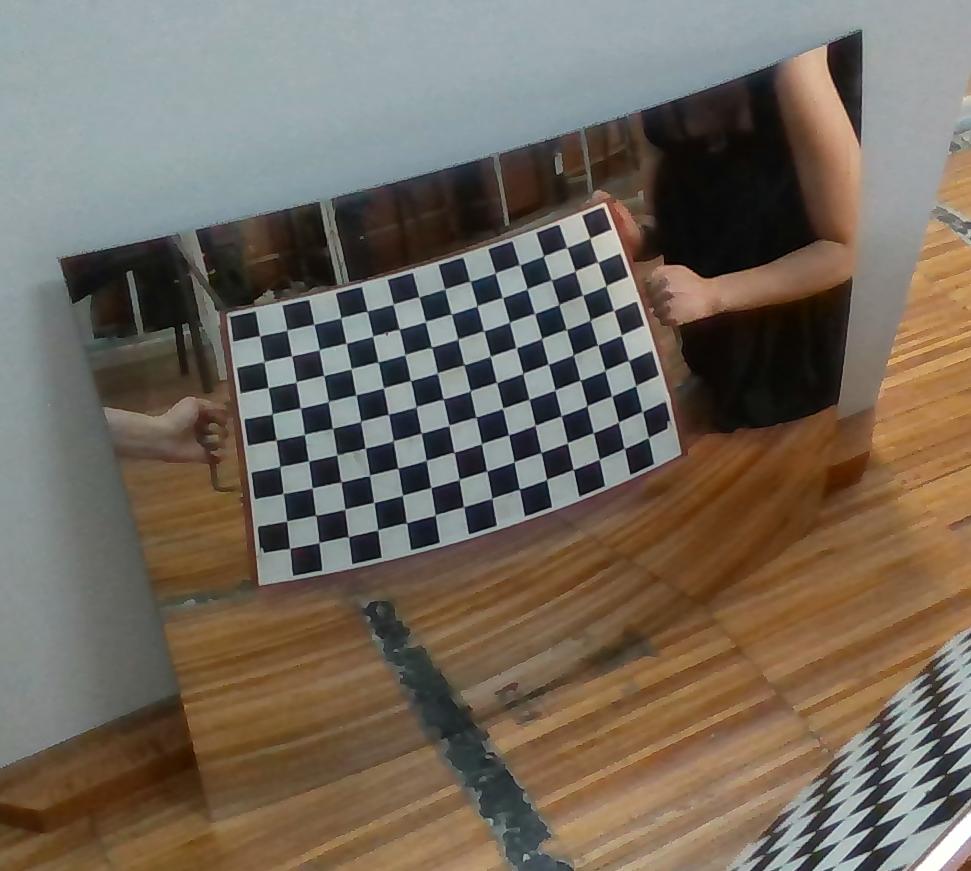}
     \end{subfigure}
     \begin{subfigure}[m]{0.23\textwidth}
     \centering
     \includegraphics[width=\textwidth]{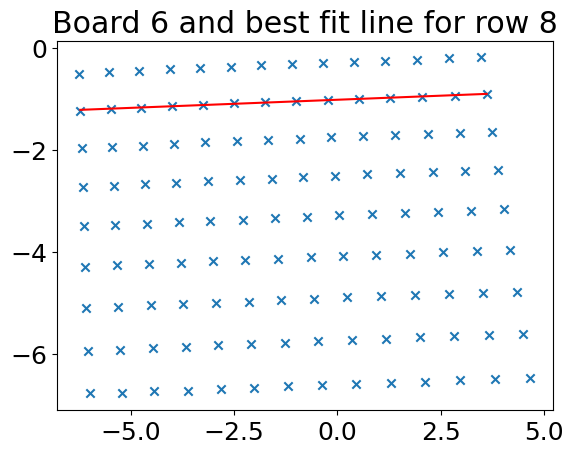}
     \end{subfigure}
     \begin{subfigure}[m]{0.23\textwidth}
     \centering
     \includegraphics[width=\textwidth]{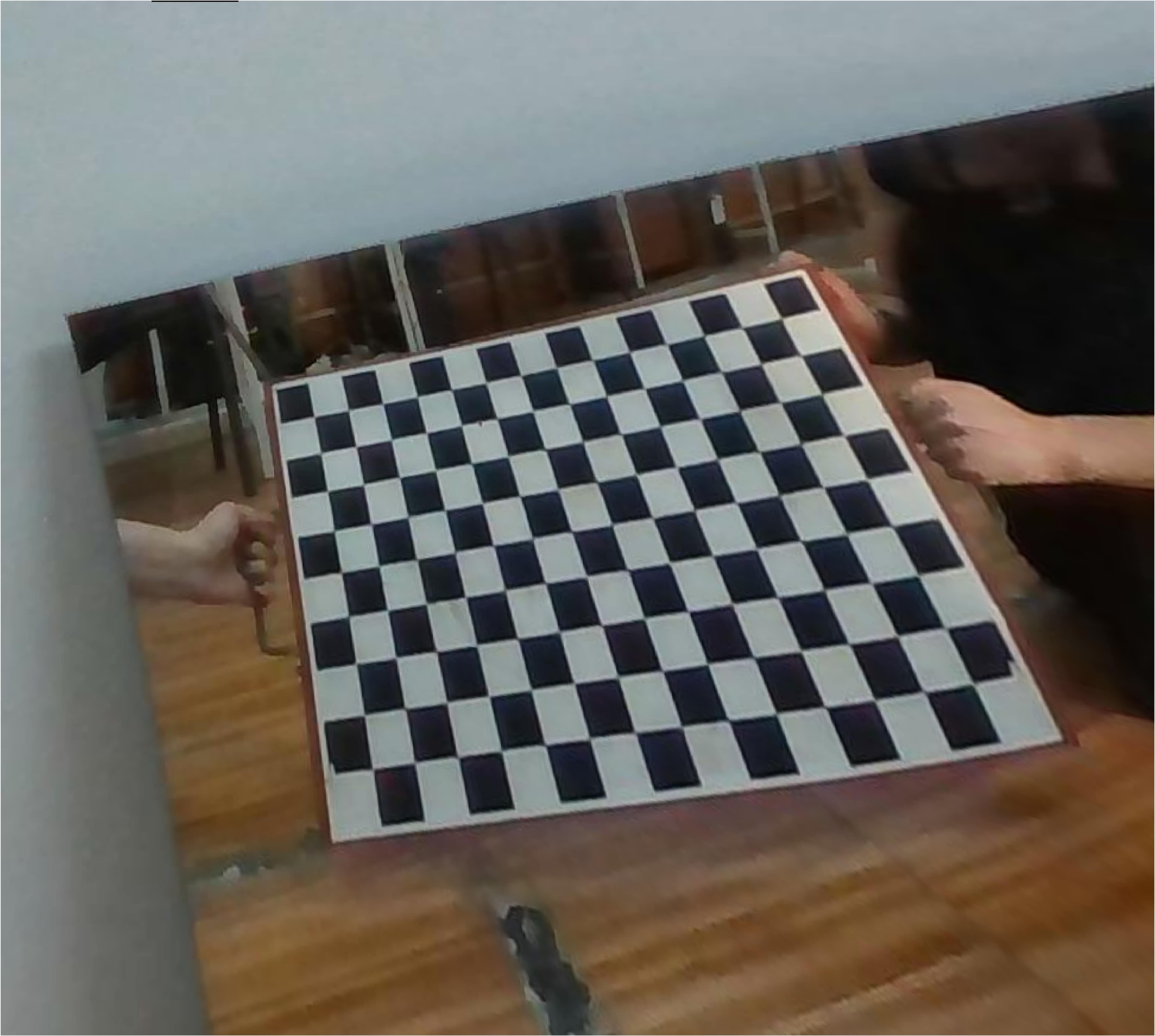}
     \end{subfigure}
    \begin{subfigure}[m]{0.23\textwidth}
    \centering
    \includegraphics[width=\textwidth]{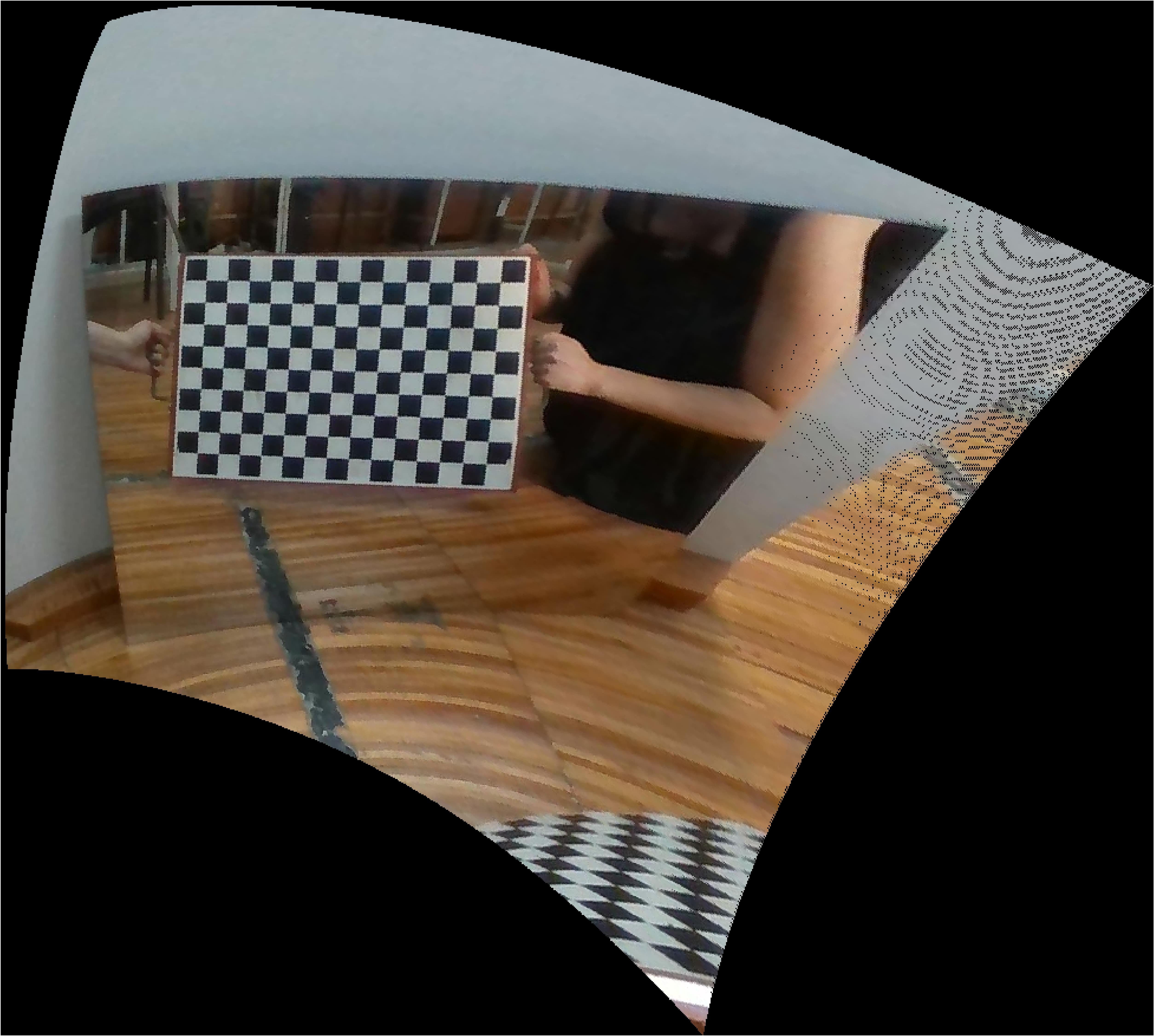}
    \end{subfigure}

    \caption{Column 1: an example of one board of the datasets of Table \ref{table:T2}. Column 2: the collinearity check on the result of the GP for that board. Column 3: undistorted image by MATLAB. Column 4: undistorted image by Gaussian processes (our method).}
    \label{fig:all_images}
\end{figure}

The last three datasets are made with real world cameras. The Webcam is an Avalue 2k Webcam and the telelens is an Apexel Telelens x2. The RealSense is of the type D415. This can be seen as catadioptric system, as the camera is pointed at a mirror with unknown curvature. The centre is relatively flat while the edges show more spherical bending. There is no mathematical model to calibrate this system. In Figure \ref{fig:all_images} we depict an example of one board out of every dataset in the first column.

\section{Results}\label{Results}
We validate our findings with three assessments. First, we investigate how well the Gaussian processes predict collineations, meaning predictions for corners of a checkerboard result in rows and columns that are straight lines. Second, we calculate the reprojection error for found 3D coordinates of corners of the used checkerboards. Finally, we demonstrate the removal of distortion.

\subsection{Collineation assumption}\label{H_Ass}

In this section, we validate our method by showing that the mapping done by our GP-based algorithm from $uv$- to $xy$-coordinates is a projective transformation where any straight line fed into the said mapping remains straight. In other words, collinearity of points is preserved. We prove the aforementioned statement both visually (Figure \ref{fig:all_images}) and quantitatively (Table \ref{table:T2}). 

Let the grid formed by the corner points of a checkerboard have several rows and columns of multiple points each. We find the best fit lines through all of these individual rows and columns. Subsequently, we calculate the perpendicular distance of each point from the corresponding best fit line. The Root Mean Square (RMS) of these perpendicular distances for each row/column is divided by the distance between the end points of the respective row/column to obtain the unitless version of this RMS value. Finally, these scaled unitless RMS values are averaged across all the boards in each dataset to get the Average Root Mean Square Collinearity Error, which is abbreviated as GP CE in the third column of Table \ref{table:T2}. For visual confirmation, we show one such best fit line for a given board in every dataset in the second column of Figure \ref{fig:all_images}.

It is worth mentioning that the positions of the corners on the virtual image plane should be interpreted by the projection of the virtual Gaussian process camera, whose location and rotation differs from the real (or Unity) camera. This depends on the image of the first checkerboard (see also Section \ref{sec:ideal}). In other words, they are not just unwarpings of the original image.

\subsection{Reprojection error}\label{Reproj}

In our method, we calculate a camera matrix $\mathbf{K}$ for our GP-camera such that for each board, the corresponding homography $\mathbf{H}$ has the form $\mathbf{H} \sim \mathbf{K} [\mathbf{r_1}\mid\mathbf{r_2}\mid\mathbf{t}]$. Furthermore, $\mathbf{K}$ has the form
\begin{equation}\label{K_sim2} 	 	
\mathbf{K} = \begin{pmatrix}f & 0 & u_c \\ 0 & f & v_c \\ 0 & 0 & 1 \\ \end{pmatrix} .
\end{equation}
When solving the overdetermined linear system of equations (see Appendix \ref{App_Zhangs_simplified}), we observe that we find a small value for the smallest singular values, meaning our method is in agreement with an algebraic point of view.

To interpret this from a geometrical point of view, we reproject the 3D coordinates of the corners to the image plane using the parameters found in the camera calibration process. Next, we compare those to the Gaussian process predictions and the found $uv$-coordinates of the corners for our method and the MATLAB Camera Calibration Toolbox (version R2023a) respectively. As done previously, we calculate the RMS error for these values and scale them by dividing by the distance between two corners, making them unitless. These errors are given in Table \ref{table:T2}. From this table, we can see that our method outperforms the MATLAB Camera Calibration Toolbox, especially for datasets with severely warped images such as the Unity with severe barrel distortion, the Unity with eccentric pincushion distortion and the RealSense with curved mirror. This is due to the fact that the MATLAB Camera Calibration Toolbox is based on a model that oversimplifies these underlying realities.

\setlength{\tabcolsep}{0.5cm}
\begin{table}[]
\centering
\caption{Collinearity Errors (CE) and Reprojection Errors (RE)}
\begin{tabular}{llll}
\hline
Dataset &    GP CE $(\times 10^{-4})$   & GP RE     &    MATLAB RE     \\ 
\hline
Unity pinhole   & 0.954         & 0.1197     & 0.1229 \\
Unity barrel    & 1.334         & 0.1410     & 1.0315 \\
Unity pincushion & 1.062        & 0.1243     & 0.3879 \\
Webcam          & 3.077         & 0.2478     & 0.2666  \\
Webcam with telelens  & 3.903   & 0.2576     & 0.2667 \\
RealSense with mirror & 9.455   & 0.4404     & 0.5376 \\
\hline
\end{tabular}
\label{table:T2}
\end{table}

\begin{figure}[t!]
     \centering
     \begin{subfigure}[m]{0.32\textwidth}
         \centering
         \includegraphics[width=\textwidth]{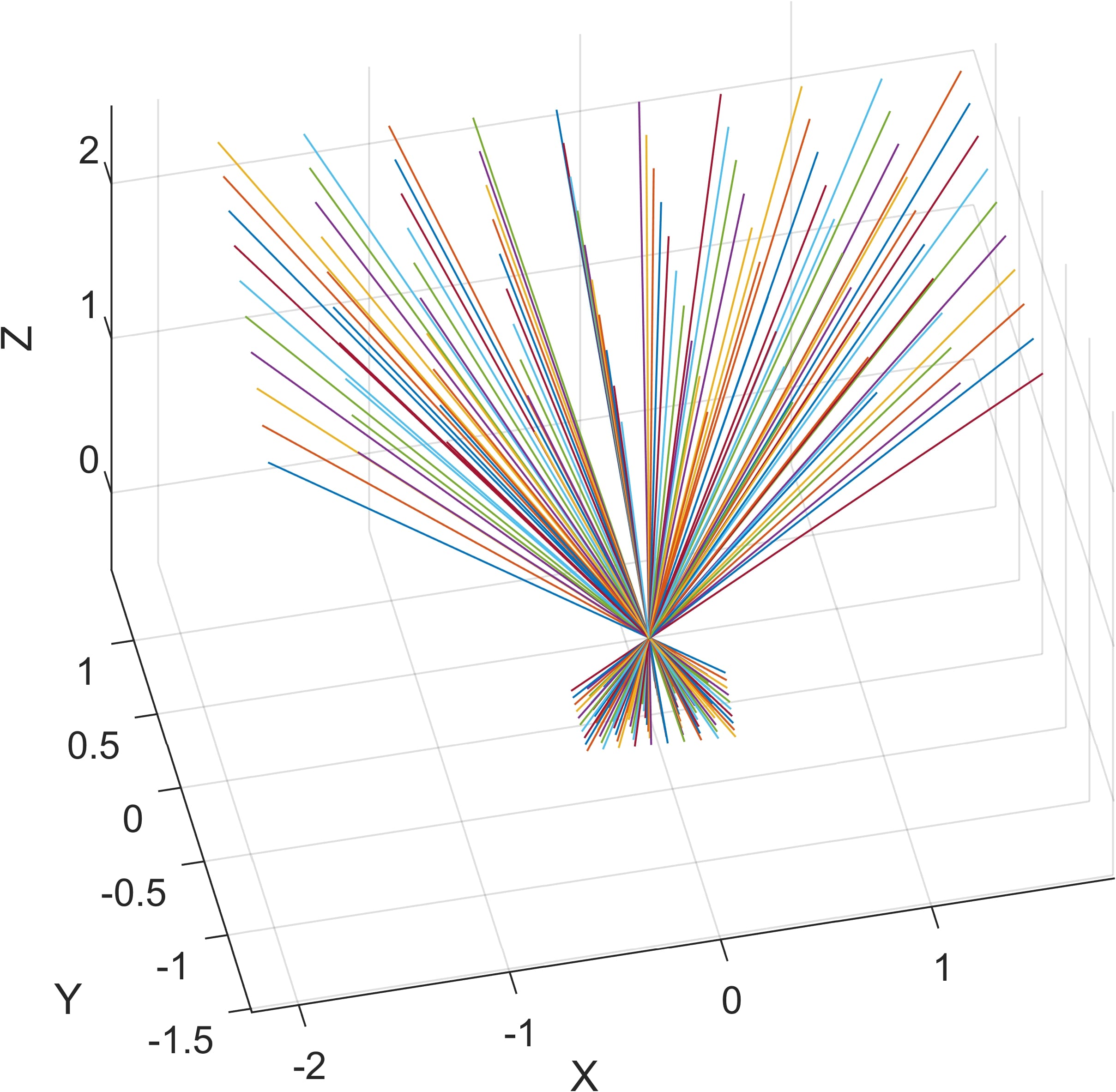}
         \caption{}
     \end{subfigure}
     \hfill
     \begin{subfigure}[m]{0.32\textwidth}
         \centering
         \includegraphics[width=\textwidth]{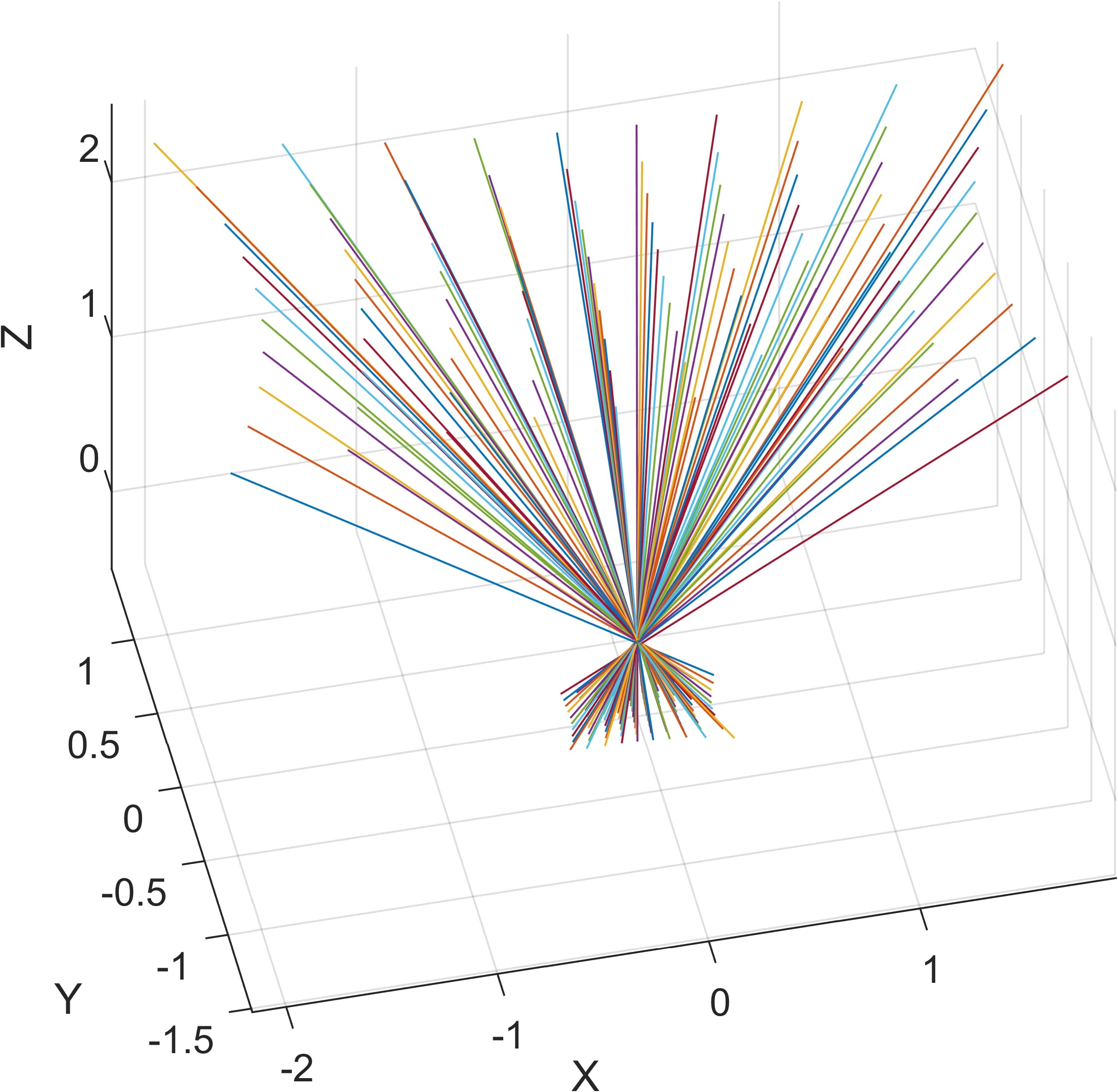}
         \caption{}
     \end{subfigure}
     \hfill
     \begin{subfigure}[m]{0.32\textwidth}
         \centering
         \includegraphics[width=\textwidth]{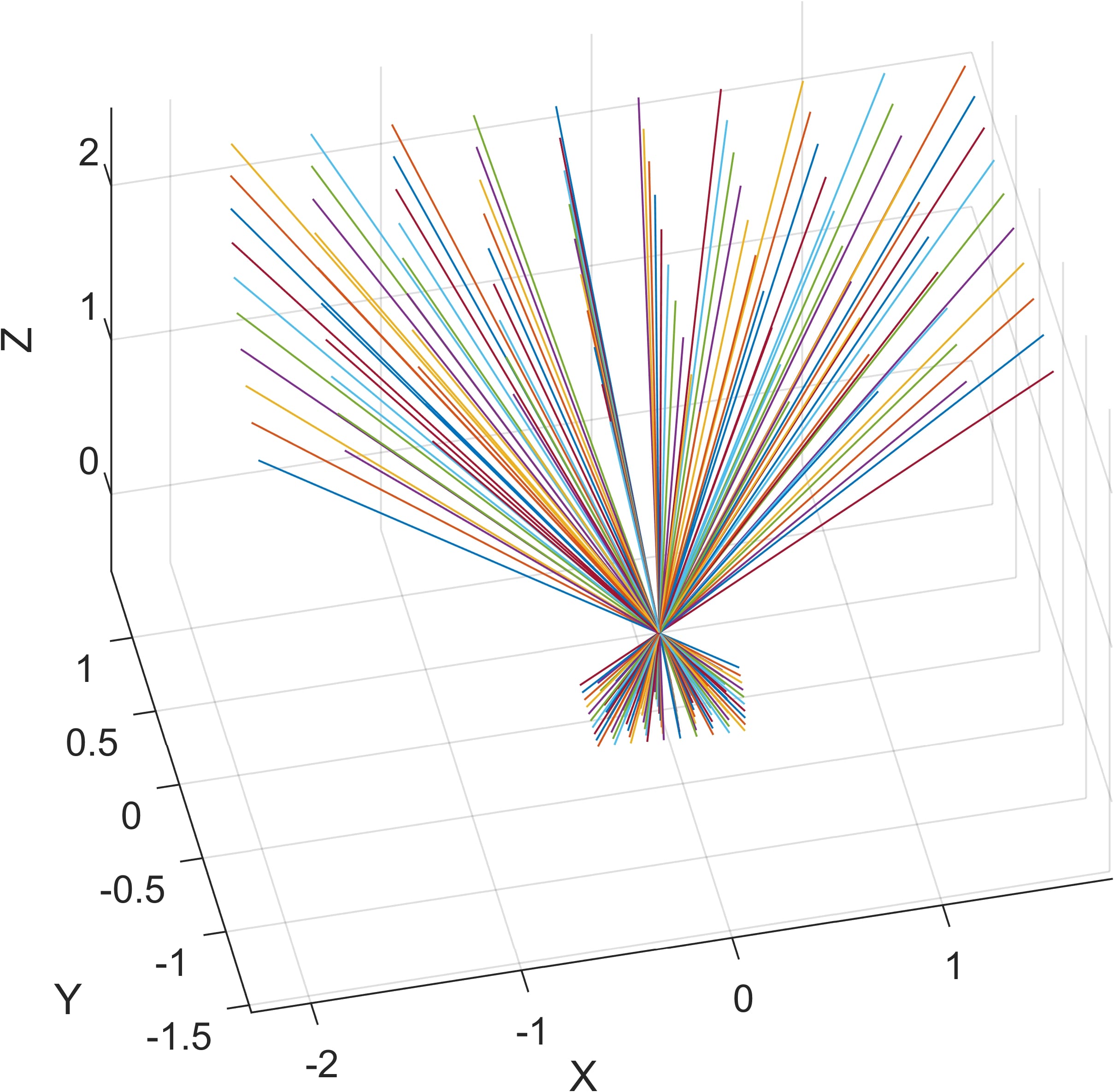} 
         \caption{}
     \end{subfigure}
    \vfill
    \begin{subfigure}[m]{0.32\textwidth}
         \centering
         \includegraphics[width=\textwidth]{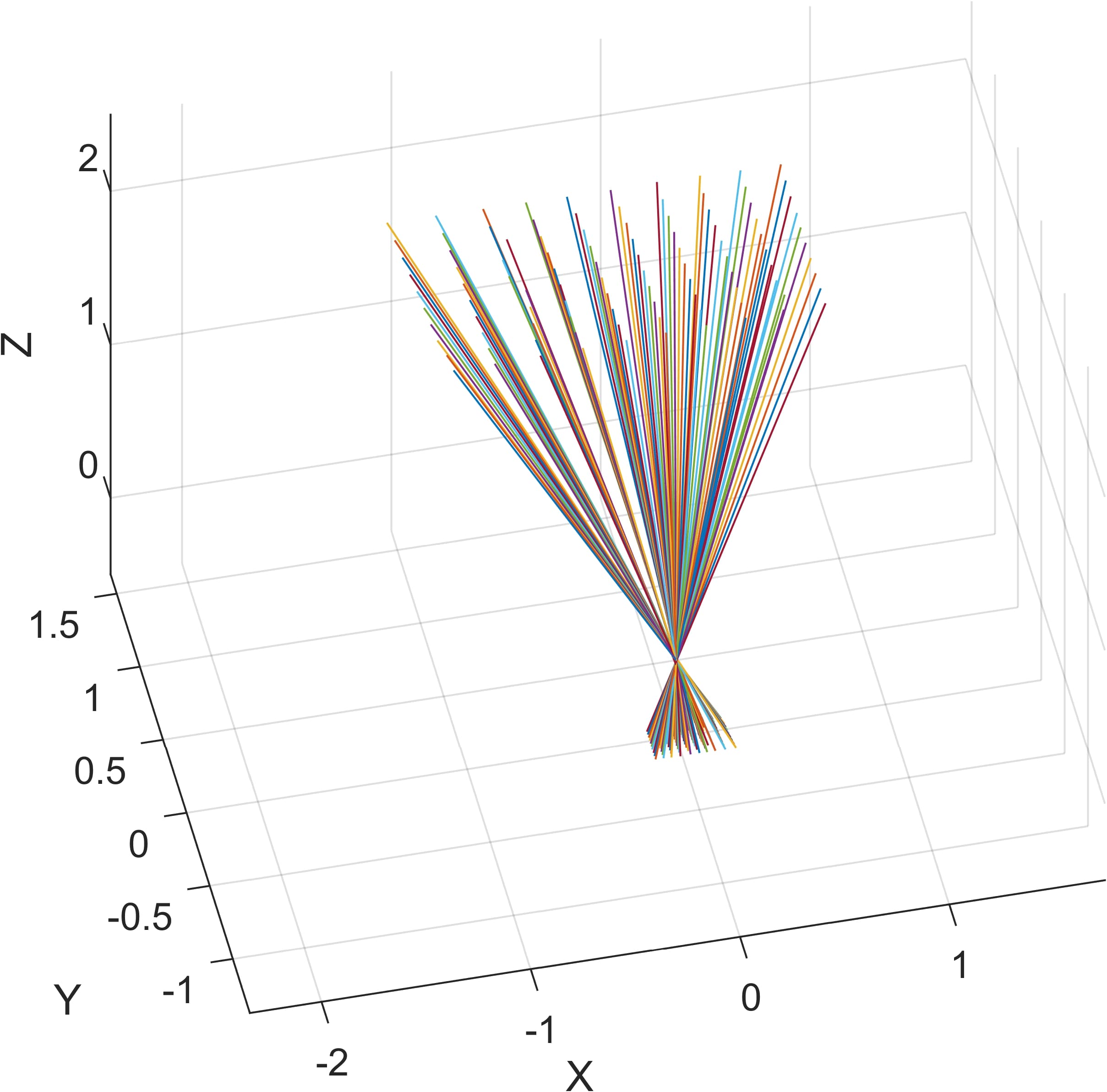}
         \caption{}
     \end{subfigure}
     \hfill
     \begin{subfigure}[m]{0.32\textwidth}
         \centering
         \includegraphics[width=\textwidth]{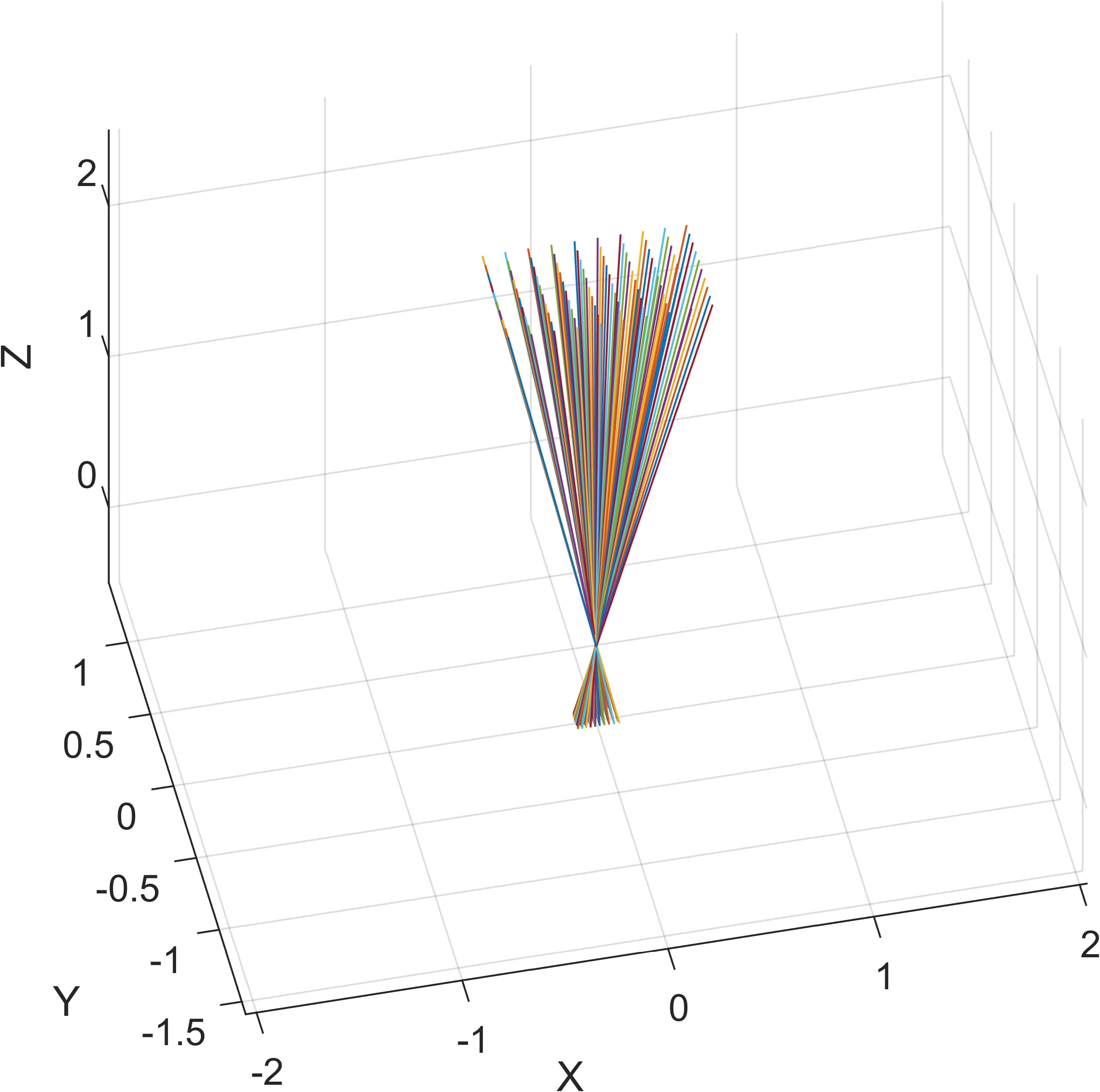}
         \caption{}
     \end{subfigure}
     \hfill
     \begin{subfigure}[m]{0.32\textwidth}
         \centering
         \includegraphics[width=\textwidth]{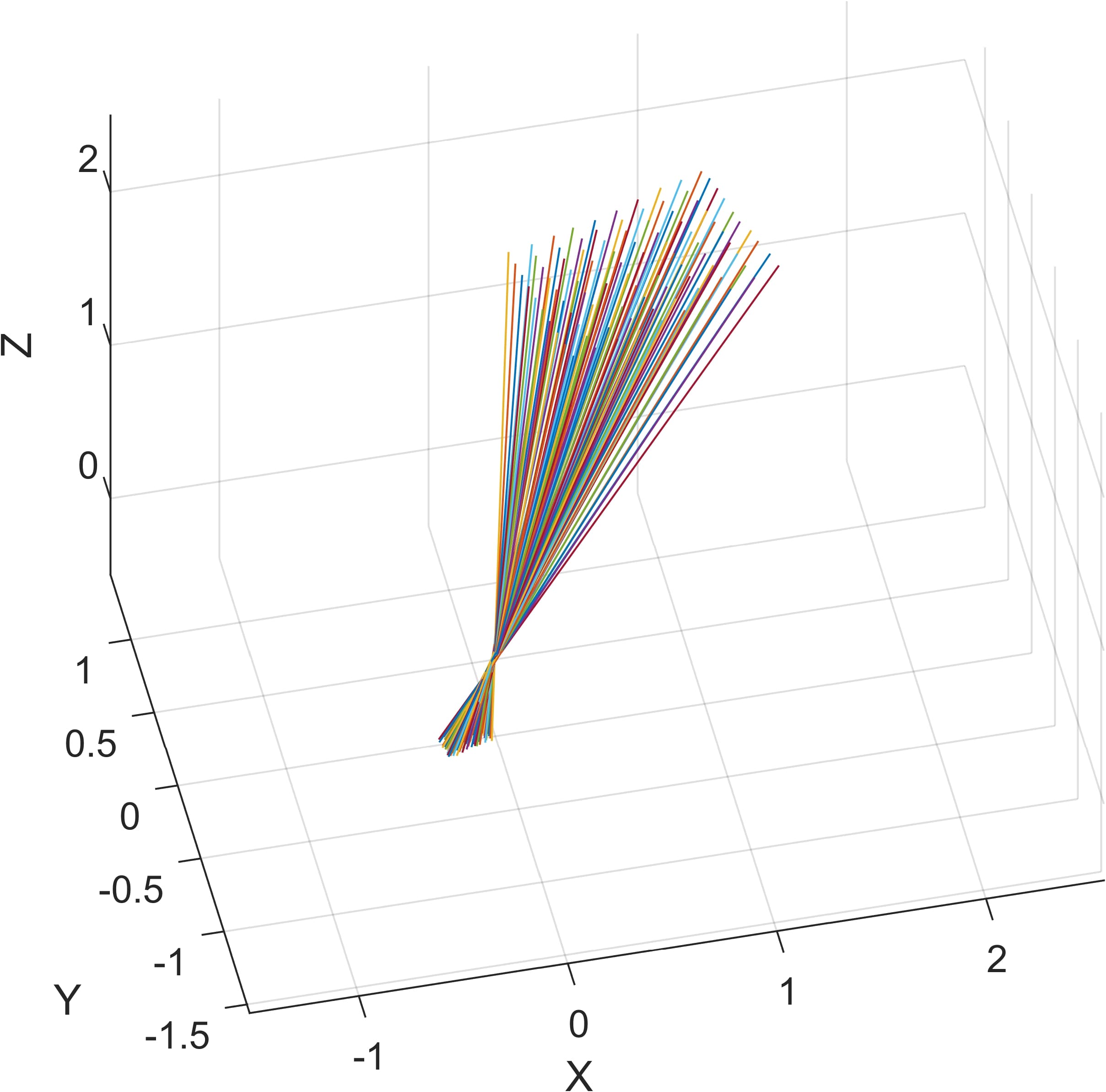}
         \caption{}
     \end{subfigure}  
    \caption{Visualisation of the pinhole results for each of the six datasets. Only (a) and (d) are actual pinhole cameras, in Unity and the real world respectively.}
    \label{fig:pinholes}
\end{figure}

We demonstrate the \textit{pinholeness} of our GP-camera further by visualising a grid of 10x10 lines that accompany a given set of pixels. For a pinhole camera, all lines intersect the optical centre, which is also the centre of the reference frame. This implies that all our pixels should correspond to straight lines going through the origin of the reference system. First, we define a 10x10 subset of pixels. We know $\mathbf{R_n}$ and $\mathbf{t_n}$ for each of the $n$ checkerboards  from the camera calibration. For each $xy$-coordinate of the $10\times10$ pixels, we calculate a corresponding 3D point on each checkerboard. Next, we group together coordinates of 3D points that belong to the same pixel. On these grouped points, we perform a least squares best fit line \cite{lesueur2014least}, including RANSAC. An alternative mechanical-inspired approach is given in \cite{penne2008mechanical}. We visualise these lines in Figure \ref{fig:pinholes}. Notice how, for instance, the barrel distortion manifests itself as a warping of the grid of lines, while retaining the pinhole model.

\subsection{Distortion removal}\label{Rem_Dis}

The trained GP predicts a new location in virtual $xy$-coordinate frame for every pixel $uv$-coordinate frame of the original distorted image. Based on those virtual pixel values, we distil a new image. The predicted coordinates are non-integer numbers, which we round to an integer value. This rounding could imply that some pixels are left empty (black). This means no original pixel is mapped to that specific virtual pixel. We solve this issue by implementing a median filter on the surrounding pixels. Alternatives to this exist, but are outside the scope of this paper. The resulting undistorted images are shown in the last column of Figure \ref{fig:all_images}. Notice how our method is better equipped to handle distortions, as it is not limited to an underlying oversimplifying distortion model. The difference is most notable for severely warped checkerboards.

\section{Discussion}\label{Discussion}

The fact that collinearity is preserved when the GPs map $uv$-coordinates of found corners to the virtual image plane proves the projective transformation between the real world (including Unity) checkerboards and their virtual images. This means there is a homography between two virtual images of two checkerboards, validating our approach. All non-linearities are captured by the GPs. The decomposition of the projective transformation for every board contains the same $\mathbf{K}$. We make use of this to calculate a line in 3D space for a given pixel. We observe that every line that corresponds to a pixel goes through the origin of the reference system of the GP-camera. This demonstrates the pinhole behaviour of our GP-camera.

An accurate corner detection algorithm is a crucial first step in our approach. Especially for the corners of the first checkerboard, on which the Gaussian processes are trained. Moreover, the checkerboard should consist of a sufficient amount of corners so that the distortion can be fully captured. 

The reference system of the GP-camera is different from the real (or Unity) camera. This is the result of how the Gaussian process is trained. All $uv$-coordinates of corners are mapped to a regular square grid. This means the first board is perpendicular to the optical axis and all rows of points are horizontal. This can be seen in the last column of Figure \ref{fig:all_images}.

From these images we can also observe that for the severe barrel distortion and the eccentric pincushion distortion (row two and three), the MATLAB method fails. Our method is more flexible and can capture these. Notice how the undistortion in the lower left region is better in the last image of row three.

Our Gaussian process predictions decrease in reliability as we move away from the region of the corners in the image of the first board. For those points, the Gaussian processes fall back on their prior, accompanied by large uncertainties. The latter can be taken into account. We retained the pixel predictions with a large uncertainty in the last image of the last row of Figure \ref{fig:all_images} for demonstration purposes. Notice that the results become meaningless far away from the first checkerboard where the GP was trained on.

A fallacy of working with Gaussian processes is that they sometimes smooth things out too much. Especially when working with a squared exponential kernel. This is an issue for example, when working with fisheye images, where a lot of corners are situated at the edge of the image and only a few are at the centre of the image. This data is in essence non-stationary. It varies more in some regions (the edges) than in others (the centre). A solution for this is to work with more corners, a different kernel or even active targets with Gray code instead of corners \cite{Sels2019}.

\section{Conclusion}\label{Conclusion}
The aim of the present research was to construct a virtual ideal pinhole camera out of a given camera (including catadioptric systems with mirrors). We showed how this is possible by means of Gaussian processes, which capture everything that makes the camera deviate from an ideal pinhole model. This includes lens distortions and imperfections. Experiments confirmed that our approach results in a pinhole camera.

Further work is required to establish the benefit of our approach in real world camera calibration and compare them to other state of the art methods. We will address this in upcoming publications, as there exists a myriad of camera systems and likewise calibration procedures.

Our model allows for a serious upgrade of many algorithms and applications that are designed in a pure projective geometry setting but with a performance that is very sensitive to non-linear lens distortions.

\subsubsection*{Funding}

The authors would like to acknowledge funding from the following PhD scholarships: BOF FFB200259, Antigoon ID 42339; UAntwerp-Faculty of Applied Engineering; 2020.06592.BD funded by FCT, Portugal and the Institute of Systems and Robotics - University of Coimbra, under project UIDB/0048/2020.

\subsubsection*{Acknowledgements} 
Conceptualization, I.D.B. and R.P.; methodology, I.D.B., S.P. and R.P; software I.D.B.; validation, S.P.; formal analysis, I.D.B., S.P. and R.P; data curation, I.D.B. and M.O.; writing original draft preparation, I.D.B. and S.P.; writing review and editing, I.D.B., S.P., M.O. and R.P; supervision, R.P.; project administration, R.P.; funding acquisition, R.P.

%
%
%
\bibliographystyle{splncs04}
\bibliography{mybibliography}

\section*{Appendix}
\appendix

\section{Zhang's method}\label{App_Zhangs}

The intrinsic camera matrix for a pinhole camera $\mathbf{K}$ can be written as
\begin{equation}\label{K} 	 	
\mathbf{K} = \begin{pmatrix}f s_x & f s_{\theta} & u_c \\ 0 & f s_y & v_c \\ 0 & 0 & 1 \\ \end{pmatrix} ,
\end{equation}
in which $f$ is the focal length, $s_x$ and $s_y$ are sensor scale factors, $s_{\theta}$ is a skew factor and $(u_c,v_c)$ is the coordinate of the image centre with respect to the image coordinate system. 
However, real world cameras and their lenses suffer from imperfections. This introduces all sorts of distortions, of which radial distortion is the most commonly implemented. Calibrating this non-ideal pinhole camera, means finding values for both $\mathbf{K}$ and $[\mathbf{R}\mid\mathbf{t}]$, and whichever distortion model is implemented.

Zhang's method is based on the images of checkerboards with known size and structure. For each position of the board, we construct a coordinate system where the $X-$ and $Y$-axis are on the board and the $Z$-axis is perpendicular to it. We assign all checkerboard corners a 3D coordinate in this system with a $Z$-component zero. This allows us to rewrite Equation \ref{P_decomposed_1}
\begin{equation}\label{DLT} 	 	
\begin{pmatrix}x\\y\\1\end{pmatrix} = \mathbf{K} [\mathbf{R}\mid\mathbf{t}] \begin{pmatrix}X\\Y\\0\\1\end{pmatrix} =  \mathbf{K} [\mathbf{r_1}\mid\mathbf{r_2}\mid\mathbf{t}] \begin{pmatrix}X\\Y\\1\end{pmatrix} ,
\end{equation}
in which $\mathbf{r_1}$ and $\mathbf{r_2}$ are the first two columns of $\mathbf{R}$. This equation shows a 2D to 2D correspondence known as a homography. This means we can write
\begin{equation}\label{H} 	 	
\begin{pmatrix}x\\y\\1\end{pmatrix} = \mathbf{H} \begin{pmatrix}X\\Y\\1\end{pmatrix}  ,
\end{equation}
with $\mathbf{H}$ the 3x3 matrix that describes the homography. This matrix is only determined up to a scalar factor, so it has eight degrees of freedom. Each point correspondence yields two equations. Therefore, four point correspondences are needed to solve for $\mathbf{H}$. In practice, we work with several more points in an overdetermined system to compensate for noise in the measurements.

From these homographies, one for every position of the checkerboard, we estimate the camera intrinsics and extrinsic parameters. From Equation \ref{DLT} and \ref{H}, we can write a decomposition for $\mathbf{H}$, up to a multiple, as
\begin{equation}\label{DLT_H} 	 	
\lambda \mathbf{H} = \lambda [\mathbf{h_1}\mid\mathbf{h_2}\mid\mathbf{h_3}] = \mathbf{K} [\mathbf{r_1}\mid\mathbf{r_2}\mid\mathbf{t}],
\end{equation}
where $\lambda$ is a scaling factor and $\mathbf{h_1}$, $\mathbf{h_2}$ and $\mathbf{h_3}$ are the columns of $\mathbf{H}$. We observe the following relationships: 
\begin{equation}\label{c_r1} 	 	
\lambda \mathbf{K^{-1}} \mathbf{h_1} = \mathbf{r_1} ,
\end{equation}
\begin{equation}\label{c_r2} 	 	
\lambda \mathbf{K^{-1}} \mathbf{h_2} = \mathbf{r_2} .
\end{equation}
Moreover, since $\mathbf{R}$ is a rotation matrix, it is orthonormal. This means $\mathbf{r_1}^T \mathbf{r_2} = 0$ and $\Vert\mathbf{r_1}\Vert =  \Vert\mathbf{r_2}\Vert$.
Combining these equations yields
\begin{equation}\label{B1} 	 	
\mathbf{h_1}^T \mathbf{K^{-T}} \mathbf{K^{-1}} \mathbf{h_2} = 0 ,
\end{equation}
\begin{equation}\label{B2} 	 	
\mathbf{h_1}^T \mathbf{K^{-T}} \mathbf{K^{-1}} \mathbf{h_1} = \mathbf{h_2}^T \mathbf{K^{-T}} \mathbf{K^{-1}} \mathbf{h_2} .
\end{equation}
These are now independent of the camera extrinsics.

We can write $\mathbf{K^{-T}} \mathbf{K^{-1}}$ as a new symmetric 3x3 matrix $\mathbf{B}$, alternatively by a 6-tuple $\mathbf{b}$. From Equations \ref{B1} and \ref{B2} we can write $\mathbf{A} \mathbf{b} = 0$, in which $\mathbf{A}$ is composed out of all known homography values of the previous step and $\mathbf{b}$ is the vector of six unknowns to solve for. For $n$ checkerboards, and thus $n$ homographies, we now have $2n$ equations. This means we need at least three checkerboard positions. Once $\mathbf{b}$ and thus $\mathbf{B}$ is found, we can calculate $\mathbf{K}$ via a Cholesky decomposition on $\mathbf{B}$. From $\mathbf{K}$, we know all camera intrinsics such as skewness, scale factor, focal length and principal point.

From Equations \ref{c_r1} and \ref{c_r2} we can determine $\mathbf{r_1}$ and $\mathbf{r_2}$. The scaling factor $\lambda$ can be found by normalising $\mathbf{r_1}$ and $\mathbf{r_2}$ to unit length. Building on the orthogonality of the rotation matrix $\mathbf{R}$, we can write
\begin{equation}\label{r3} 	 	
\mathbf{r_3} = \mathbf{r_1} \times \mathbf{r_2} .
\end{equation}
Lastly, we find 
\begin{equation}\label{t} 	 	
\mathbf{t} = \lambda \mathbf{K}^{-1} \mathbf{h_3} .
\end{equation}

Up until this point, we have assumed an ideal pinhole camera model. MATLAB and OpenCV use this as a first step in an iterative process in which they introduce extra intrinsic camera parameters to account for image distortion. After convergence, a compromise is found for all camera parameters. 

\section{Simplified Zhang's method}\label{App_Zhangs_simplified}
In this work, we construct an ideal virtual GP-camera. The Gaussian processes capture all distortions and other imperfections in a pre-processing step. This means that the images of the checkerboards on the virtual image plane are projections of a perfect checkerboard, up to noise. This allows us to simplify Zhang's method as follows.

First, since there is no skewness in the virtual image plane and all virtual pixels are squares, we can rewrite the intrinsic camera matrix $\mathbf{K}$ as
\begin{equation}\label{K_sim} 	 	
\mathbf{K} = \begin{pmatrix}f & 0 & u_c \\ 0 & f & v_c \\ 0 & 0 & 1 \\ \end{pmatrix} .
\end{equation}
This results in
\begin{equation}\label{B_sim} 	 	
\mathbf{B} = \mathbf{K^{-T}} \mathbf{K^{-1}} = 
\begin{pmatrix}\frac{1}{f^2} & 0 & \frac{-u_c}{f^2} \\ 0 & \frac{1}{f^2} & \frac{-v_c}{f^2} \\ 0 & 0 & \frac{u_c}{f^2} + \frac{v_c}{f^2} + 1 \\ \end{pmatrix} .
\end{equation}

The rest of the procedure is similar to Zhang's method. We combine Equations \ref{B1}, \ref{B2} and \ref{B_sim} into the system $\mathbf{A} \mathbf{b} = 0$. The vector $\mathbf{b}$ is now the vector of four unknowns to solve for, instead of six. For $n$ checkerboards, and thus $n$ homographies, we still have $2n$ equations. This means we need at least two checkerboard positions to be able to solve this, instead of three. As before, more positions provide more equations, which are solved via Singular Value Decomposition (SVD). Notice that the form of Equation \ref{B_sim} is such that we do not have to perform the Cholesky decomposition anymore.

Second, there is no need for a distortion model, nor for a converging iterative process. The camera calibration is reduced to a one-step analytical calculation.

\end{document}